\documentclass[journal]{IEEEtran}
\usepackage[linesnumbered,ruled]{algorithm2e}
\usepackage{graphicx,amssymb,mathrsfs,amsmath,array,color,amsthm}
\usepackage{subfigure}
\usepackage{multirow}
\usepackage{booktabs}
\usepackage{mathtools}
\usepackage{bm}
\usepackage{cite}
\usepackage[colorlinks,urlcolor=blue,driverfallback=dvipdfm]{hyperref}

\usepackage{supertabular}
\usepackage{pifont}
\usepackage{url}
\usepackage{algpseudocode}
\usepackage{listings}
\usepackage{xcolor}
\usepackage[utf8]{inputenc}

\begin{document}
\title{Foundation Models in Remote Sensing: Evolving from Unimodality to Multimodality}

\author{Danfeng Hong,~\IEEEmembership{Senior Member,~IEEE,}
        Chenyu Li,
        Xuyang Li,
        Gustau~Camps-Valls,~\IEEEmembership{Fellow,~IEEE,}
        and Jocelyn Chanussot,~\IEEEmembership{Fellow,~IEEE}

\thanks{This work was supported by the National Natural Science Foundation of China under Grant 42271350 and by the International Partnership Program of the Chinese Academy of Sciences under Grant No.313GJHZ2023066FN. GCV gratefully acknowledges the support from the European Research Council (ERC) through the USMILE project (grant agreement 855187) and the European Commission through the HORIZON projects ELIAS (grant agreement 101120237) and ELLIOT (grant agreement101214398).}
\thanks{D. Hong is with the School of Automation, Southeast University, Nanjing 211189, China. (e-mail: danfeng.hong@seu.edu.cn)}
\thanks{C. Li is with the School of Mathematics, Southeast University, Nanjing 211189, China. (e-mail: chenyuli.erh@gmail.com)}
\thanks{X. Li is with the Aerospace Information Research Institute, Chinese Academy of Sciences, Beijing 100094, China, and the School of Electronic, Electrical and Communication Engineering, University of Chinese Academy of Sciences, Beijing 100049, China.}
\thanks{G. Camps-Valls is with the Image Processing Laboratory (IPL), Universitat de València, Paterna, València 46980, Spain. (e-mail: gustau.camps@uv.es)}
\thanks{J. Chanussot is with Univ. Grenoble Alpes, INRIA, CNRS, Grenoble INP, LJK, Grenoble 38000, France. (e-mail: jocelyn.chanussot@grenoble-inp.fr)}
}

\markboth{}
{Shell \MakeLowercase{\textit{et al.}}: }

\maketitle
\begin{abstract}
Remote sensing (RS) techniques are increasingly crucial for deepening our understanding of the planet. As the volume and diversity of RS data continue to grow exponentially, there is an urgent need for advanced data modeling and understanding capabilities to manage and interpret these vast datasets effectively. Foundation models present significant new growth opportunities and immense potential to revolutionize the RS field. In this paper, we conduct a comprehensive technical survey on foundation models in RS, offering a brand-new perspective by exploring their evolution from unimodality to multimodality. We hope this work serves as a valuable entry point for researchers interested in both foundation models and RS and helps them launch new projects or explore new research topics in this rapidly evolving area. This survey addresses the following three key questions:
\begin{itemize}
    \item What are foundation models in RS?
    \item Why are foundation models needed in RS?
    \item How can we effectively guide junior researchers in gaining a comprehensive and practical understanding of foundation models in RS applications?
\end{itemize}
More specifically, we begin by outlining the background and motivation, emphasizing the importance of foundation models in RS. We then review existing foundation models in RS, systematically categorizing them into unimodal and multimodal approaches. Additionally, we provide a tutorial-like section to guide researchers, especially beginners, on how to train foundation models in RS and apply them to real-world tasks. The survey aims to equip researchers in RS with a deeper and more efficient understanding of foundation models, enabling them to get started easily and effectively apply these models across various RS applications.
\end{abstract}
\graphicspath{{figures/}}


\section{Introduction}
\IEEEPARstart{R}{ecent} advancements in Earth Observation (EO) technologies, particularly Remote Sensing (RS), have revolutionized our ability to monitor various aspects of the Earth’s surface and environment, such as land features, subsurface conditions, air and water quality, and the health of living organisms \cite{hong2024multimodal}. EO technologies are essential tools for gathering critical information about the Earth’s physical properties, environmental conditions, and dynamic changes. Monitoring these aspects from space allows for more accurate and timely decision-making, improving natural resource management, disaster preparedness, and environmental protection.

These capabilities advance scientific understanding of Earth systems and play a crucial role in addressing global challenges, including climate change, biodiversity loss, and sustainable development. RS data, which offers a unique vantage point for studying the Earth at various scales, provides insights that are difficult or impossible to obtain through traditional ground-based observation methods. The integration of these insights can drive actions that improve human well-being, promote societal stability, and ensure environmental sustainability.

\subsection{The Rising Role of Foundation Models in Remote Sensing}
However, the rapid growth in the volume and diversity of RS data has outpaced the capabilities of traditional models, necessitating the development of more sophisticated approaches for processing and analyzing high-dimensional data. As RS datasets become more complex, containing satellite imagery, spectral data, temporal sequences, and sensor modalities like LiDAR, traditional methods struggle to capture the richness of information in these data sources. This highlights the urgent need for next-generation AI models capable of handling multimodal, spatiotemporal, and high-dimensional data.

The advent of foundation models provides an effective solution to this challenge \cite{li2024interpretable}. These large-scale models, which can be pre-trained on vast, diverse datasets and fine-tuned for specific tasks, offer significant potential for RS applications. They can leverage the scale and complexity of RS data to develop transferable representations applicable to a wide range of geospatial tasks, such as land cover classification, climate monitoring, disaster management, and agriculture. The versatility of foundation models is particularly valuable in scenarios where labeled data are scarce or costly to obtain, commonly in RS, where pixel-level annotations for large areas are labor-intensive and resource-consuming. The self-supervised pretraining and fine-tuning approach of foundation models allows large volumes of unlabeled data to learn universal features that can be applied across various RS tasks. This framework enhances the scalability and efficiency of RS analysis, making it possible to better capture the complex, multimodal, and dynamic nature of EO data \cite{Bodnar2024a, chen2023foundation, rolf2024mission, li2025urbansam}.

\begin{table*}[!t]
    \centering
    \caption{A list of review papers on foundation models in RS, geoscience, and EO, detailing their titles, publication years, journals, descriptions, and reference links.}
    \resizebox{1\textwidth}{!}{
    \begin{tabular}{m{5.5cm}|c|c|m{10cm}|c}
        \toprule[1.5pt]
        \centering Title & Year & Journal & \centering Description & Ref.\\
        \hline
        \addlinespace[1.5pt]
        Brain-Inspired Remote Sensing Foundation Models and Open Problems: A Comprehensive Survey & 2023 & \footnotesize IEEE JSTARS & This review examines brain-inspired foundational models in RS, highlighting their applications, challenges, and advantages in complex tasks. & \cite{jiao2023brain} \\
        On the Promises and Challenges of Multimodal Foundation Models for Geographical, Environmental, Agricultural, and Urban Planning Applications & 2023 & \footnotesize Arxiv & This review highlights the potential, challenges, and flexibility of multimodal foundation models in geography, environment, agriculture, and urban planning through case studies. & \cite{tan2023promises} \\
        Vision-Language Models in Remote Sensing: Current Progress and Future Trends & 2024 & \footnotesize IEEE GRSM & This review explores vision-language models in RS, highlighting their progress, applications, and future trends. & \cite{li2024vision} \\
        On the Foundations of Earth and Climate Foundation Models & 2024 & \footnotesize Arxiv & This review explores the theory behind Earth and climate models, proposing a foundational framework for Earth and climate sciences. & \cite{zhu2024foundations}\\
        Towards Vision-Language Geo-Foundation Model: A Survey & 2024 & \footnotesize Arxiv & This review introduces geographic vision-language models, emphasizing geospatial priors and their multimodal fusion capabilities in geographic information processing. & \cite{zhou2024towards}\\
        AI Foundation Models in Remote Sensing: A Survey & 2024 & \footnotesize Arxiv & This review summarizes AI foundational models in RS, analyzing their applications and challenges across different tasks. & \cite{lu2024ai}\\
        Foundation model for generalist remote sensing intelligence: Potentials and prospects & 2024 & \footnotesize Science Bulletin & This review envisions universal RS models with a ``one model, multiple tasks'' architecture to tackle task fragmentation and explores bottlenecks in high-resolution image processing. & \cite{zhang2024foundation}\\
        Foundation Models for Remote Sensing and Earth Observation: A Survey & 2024 & \footnotesize Arxiv & This survey reviews RS foundation models, covering their motivation, key concepts, existing studies, benchmarks, challenges, and future directions. & \cite{xiao2024foundation}\\
        When Geoscience Meets Foundation Models: Toward a general geoscience artificial intelligence system & 2024 & \footnotesize IEEE GRSM & This review explores the potential of geoscience foundation models in advancing Earth system modeling, highlighting their advantages, construction techniques, recent advancements, and future challenges. & \cite{zhang2024geoscience}\\
        Advancements in Vision-Language Models for Remote Sensing: Datasets, Capabilities, and Enhancement Techniques & 2025 & \footnotesize RS & This review covers advancements in vision-language models, introduces RS-optimized multimodal datasets, and shifts models from descriptive analysis to decision support. & \cite{tao2025advancements}\\
        When Remote Sensing Meets Foundation Model: A Survey and Beyond & 2025 & \footnotesize RS & This review examines the potential and challenges of foundation models for RS, highlighting recent advancements, open issues, and future directions to bridge the gap between natural and RS images. & \cite{huo2025remote}\\
        {Foundation Models in Remote Sensing: Evolving from Unimodality to Multimodality (ours)} & {2025} & {--} & {This review provides a comprehensive survey of foundation models in RS, uniquely framing their evolution from unimodality to multimodality with emphasis on tutorial-style practical guidance for researchers.}& {--}\\
        \bottomrule[1.5pt]
    \end{tabular}
    }
    \label{tab:review_paper}
\end{table*}

\subsection{Challenges and Difficulties}
Despite these promising advancements, the application of foundation models in RS remains relatively underexplored. Although these models hold great potential, several barriers hinder their adoption and integration within the field, which we outline below, together with currently emerging solutions.
\begin{itemize}
    \item  The first challenge is the lack of awareness and understanding of foundation models among many researchers working in RS, EO, and geosciences. Most researchers in these domains are unfamiliar with the underlying principles of foundation models, such as self-supervised learning and transfer learning, or they lack the expertise to apply these advanced techniques in their work. This knowledge gap creates a significant barrier to the successful application of foundation models in practical RS tasks, as researchers may struggle to understand how these models work and how to adapt them to their specific needs. To mitigate this, recent tutorial-style papers and benchmark studies (e.g., BigEarthNet \cite{sumbul2019bigearthnet}, SpectralGPT\cite{hong2024spectralgpt}) have started to provide hands-on resources that help RS researchers understand pretraining and fine-tuning strategies in practice.
    \item Secondly, {although some surveys have attempted to organize and categorize RS foundation models from certain perspectives (as summarized in Table \ref{tab:review_paper}), a comprehensive, category-clear, and tutorial-oriented taxonomy of these models is still lacking.} The sheer number of existing foundation models, some of which are only incrementally different from one another, has led to confusion and uncertainty among researchers. Without a standardized framework for evaluating and categorizing these models, it becomes difficult for researchers to choose the most appropriate model for their specific application or to compare the performance of different models across various tasks. This lack of clarity exacerbates the challenges faced by the RS community in leveraging the full potential of foundation models. {In response, community-driven efforts have emerged, such as model repositories on HuggingFace dedicated to RS and standardized evaluation datasets, such as SEN12MS \cite{schmitt2019sen12ms} or C2Seg \cite{hong2023cross}, which provide more transparent benchmarks for model comparison.}
    \item Finally, there is a significant uncertainty among RS researchers about the technical usage guidelines for foundation models. Even if these models are available, many researchers remain unclear about how to fine-tune or adapt them to their specific applications. Questions about model performance, scalability, and robustness in real-world RS settings remain largely unanswered. Without proper guidelines and evaluation benchmarks, researchers may hesitate to adopt these models, fearing that their performance may not meet the needs of practical applications. Additionally, the absence of standardized protocols for integrating foundation models with existing RS systems complicates their widespread adoption and deployment. {Nonetheless, several recent works demonstrate practical adaptation pipelines, e.g., fine-tuning pretrained models for crop classification, disaster monitoring, or flood mapping. These offer concrete guidance on how to integrate foundation models into real-world RS tasks.}
\end{itemize}

These challenges significantly hinder the development and application of foundation models in RS. As the volume and complexity of RS data continue to grow, it is crucial to address these barriers to unlock the full potential of foundation models for EO. To that end, this survey aims to provide a comprehensive and accessible resource that clarifies the core concepts of foundation models and addresses the key issues obstructing their integration into RS.

\subsection{Comparison with Existing Reviews}
Recently, several reviews have explored foundation models in RS, geoscience, and EO, though the field remains relatively underexamined. Table \ref{tab:review_paper} summarizes these works by publication year.

Jiao et al. (2023) present a survey on brain-inspired foundation models in RS, discussing their unique advantages in handling complex tasks and analyzing the challenges associated with their implementation \cite{jiao2023brain}. Around the same time, Tan et al. (2023) explore the promises and challenges of multimodal foundation models across geography, environment, agriculture, and urban planning, emphasizing their adaptability through case studies \cite{tan2023promises}.

Li et al. (2024) focus on vision-language models (VLMs) in RS, providing an overview of their recent advancements, applications, and future directions \cite{li2024vision}. Meanwhile, Zhu et al. (2024) shift attention to Earth and climate foundation models, discussing their theoretical foundations and proposing a framework for their development in Earth sciences \cite{zhu2024foundations}. Zhou et al. (2024) introduce the concept of geographic vision-language foundation models, emphasizing the integration of geospatial priors and multimodal learning for geographic information processing \cite{zhou2024towards}.

Lu et al. (2024) present a broad survey on AI foundation models in RS, systematically analyzing their applications and challenges across different RS tasks \cite{lu2024ai}. Zhang et al. (2024) take a different approach by envisioning a universal RS foundation model designed for generalist intelligence, exploring the ``one model, multiple tasks'' paradigm and its potential to address task fragmentation \cite{zhang2024foundation}. Xiao et al. (2024) offer a structured review of RS foundation models, covering their motivation, key methodologies, existing benchmarks, and unresolved challenges \cite{xiao2024foundation}.

Zhang et al. (2024) extend the discussion to geoscience foundation models, exploring their role in Earth system modeling and identifying key construction techniques, advancements, and future research directions \cite{zhang2024geoscience}. Moving into 2025, Tao et al. (2025) delve into recent progress in vision-language models for RS, highlighting new multimodal datasets optimized for RS tasks and shifting the focus of these models from descriptive analysis to decision support \cite{tao2025advancements}. Lastly, another (2025) survey examines the potential and limitations of foundation models in RS, focusing on bridging the gap between natural and RS images while identifying key open challenges and future research directions \cite{huo2025remote}.

While these reviews offer valuable insights into foundation models in RS, they primarily focus on specific subdomains, methodologies, or applications. In contrast, our review work takes a broader perspective on the transition from unimodality to multimodality. We highlight key challenges and open problems in RS, clarify fundamental concepts and learning paradigms, explore emerging trends, and provide practical guidance with a strong emphasis on the future development of RS foundation models.

\begin{figure*}[!t]
   \centering
		\includegraphics[width=1.0\textwidth]{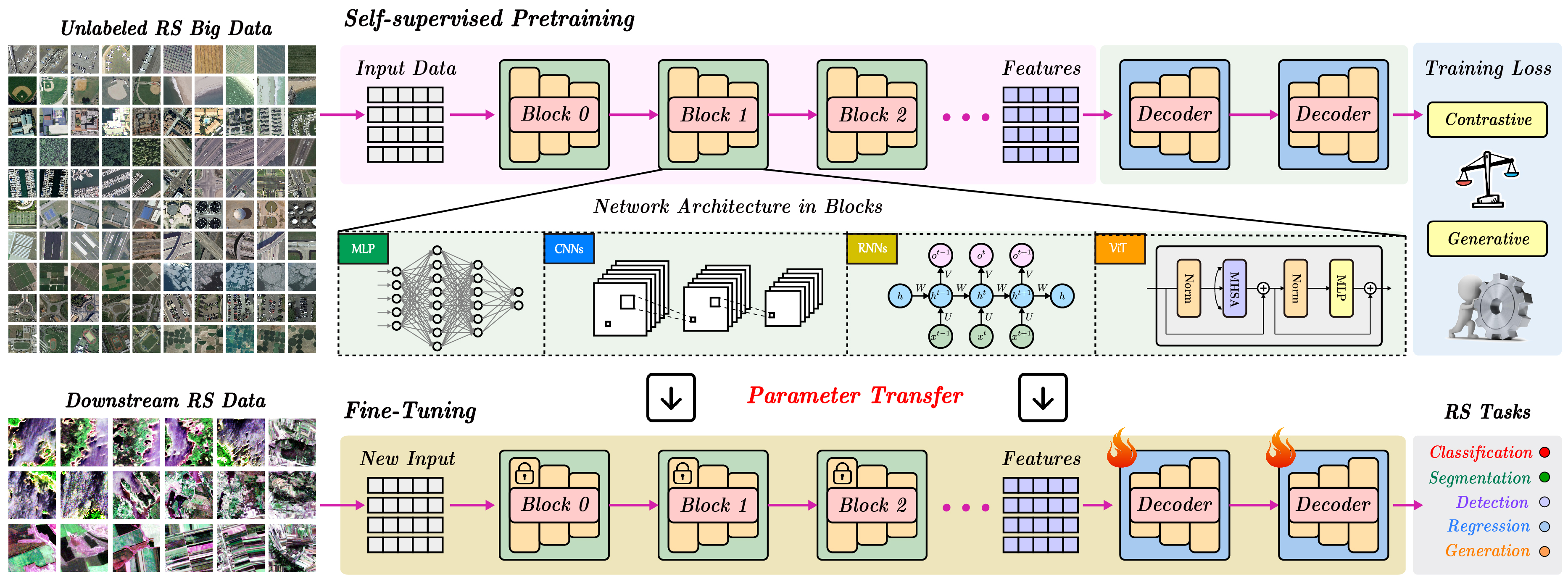}
    \caption{A general learning paradigm for foundation models in RS: self-supervised pretraining $+$ fine-tuning. {For illustration, consider a vision-related example: the model is first pretrained on large amounts of unlabeled RS data, using backbone architectures such as MLPs, CNNs, RNNs, or ViTs. Contrastive and generative (reconstruction) losses are commonly employed during pretraining. The pretrained weights are then transferred to downstream RS tasks, where they can either remain frozen or be fine-tuned (typically with emphasis on decoder weights), enabling applications such as classification, segmentation, detection, regression, and generation.}}
\label{fig:paradigm}
\end{figure*}

\subsection{Goals and Contributions}
The main goal of this paper is to create a clear pathway for RS researchers and practitioners to understand and effectively apply foundation models. We introduce these models, explain their unique learning paradigms, and demonstrate their practical applications within the context of RS. This survey emphasizes an intuitive, user-friendly approach to facilitate the broader adoption of foundation models in RS research and applications.

The contributions of this paper can be summarized as three main objectives:
\begin{itemize}
    \item We aim to provide a detailed and accessible survey of foundation models specifically tailored to the RS field. While foundation models have gained significant attention in various domains, such as natural language processing and computer vision, their application and potential in RS remain relatively underexplored. This survey is designed to bridge that gap, yielding an in-depth overview of the core principles behind foundation models, how they operate in the context of RS, and how they compare to traditional methods of RS data processing and analysis. By clearly explaining their learning paradigms, we aim to help researchers, students, and engineers in RS better understand the fundamental differences between foundation models and more conventional approaches.
    \item One of the most notable trends in the application of foundation models to RS is their evolution from unimodal models (which handle one type of data) to multimodal models (which integrate and analyze multiple types of data, such as optical, LiDAR, and SAR). This evolution represents a fundamental shift in how RS data is processed and analyzed, moving toward more holistic, integrated approaches. In this survey, we systematically review the key works in this area, providing a fresh and comprehensive perspective on how foundation models have evolved to tackle the challenges of multimodal data integration in RS. By highlighting this shift, we aim to offer readers an accessible, intuitive understanding of the increasing complexity and capability of foundation models in RS.
    \item Despite the growing awareness of foundation models within the RS community, many researchers and practitioners still struggle to understand how to effectively apply these models in real-world RS tasks. To address this, we provide a dedicated section that serves as a practical tutorial for users looking to incorporate pretrained foundation models into their work. This tutorial will walk readers through the necessary steps to apply these models, from fine-tuning pretrained models on specific RS tasks to integrating them into existing RS workflows. {By providing tutorial-like guidance, we aim to help researchers and practitioners effectively utilize RS foundation models in their respective sectors, enabling them to address the increasingly complex challenges of next-generation EO.}
\end{itemize}

\subsection{Paper Organization}
The remainder of this paper is organized as follows: Section II introduces the concept of foundation models in RS, providing a concise overview of how these models function within the RS domain. Section III reviews the evolution of foundation models in RS, tracing the shift from unimodal to multimodal models and highlighting key advancements. In Section IV, we offer a tutorial-style guide to applying pretrained foundation models in RS, helping users navigate the practical aspects of using these models for specific tasks. Finally, Section V discusses the remaining challenges facing foundation models in RS, explores potential solutions, and outlines future research directions to further improve the effectiveness and application of these models in the field. By addressing both the theoretical and practical aspects of foundation models, this survey aims to provide a valuable resource for the RS community, helping to accelerate the adoption and application of these powerful tools in EO.

\section{Foundation models}
This section provides a concise overview of foundation models from a conceptual perspective and introduces their new learning paradigm. It addresses the questions, ``What are foundation models?'' and ``How does their learning paradigm differ from previous ones?''

\subsection{A Brief Understanding of Foundation Models}

\subsubsection{\textbf{A Paradigm Shift in Machine Learning}}
Foundation models are characterized by their large scale and pretraining on vast, diverse datasets to learn generalized representations. Unlike traditional models that are typically designed for specific tasks and require extensive domain-specific training, foundation models are built to generalize across a wide range of tasks and domains. Their architecture, often based on transformer models or similar deep learning frameworks, enables them to capture complex patterns and relationships within data. This generalization capability is a result of extensive parameterization and the ability to learn from heterogeneous data sources, allowing these models to be fine-tuned with minimal additional data for specific downstream applications.

One of the most significant advantages of foundation models is their scalability. They are capable of handling large datasets and complex data structures, making them suitable for a variety of domains, from natural language processing (e.g., GPT, BERT) to computer vision (e.g., CLIP, DINO). Additionally, their transferability enables them to adapt to new tasks with limited domain-specific adaptation, drastically reducing the need for large labeled datasets. This is particularly beneficial in fields where labeled data is scarce or expensive to obtain. Furthermore, their ability to process multimodal data, such as text, images, audio, and more, allows for a more comprehensive understanding and representation of complex phenomena.

\begin{table*}[!t]
\centering
\setlength{\tabcolsep}{5pt}
\renewcommand{\arraystretch}{1.1}
\caption{Representative pretraining datasets for remote sensing foundation models.}
\label{tab:rs_datasets}
\begin{tabular}{lcccc}
\toprule
\textbf{Name}  & \textbf{Domain} & \textbf{Modality} & \textbf{Resolution} & \textbf{Image Number} \\
\midrule
SeCo               & General Scenes   & Spectral                                  & 10\,m              & About 1 million \\
SSL4EO-S12         & Land Cover       & Spectral, SAR                    & 10--60\,m          & 1,004,316 \\
fMoW               & Classification   & RGB, Spectral                           & 0.3--10\,m         & 1,047,691 \\
MillionAID        & Classification   & RGB                                          & 0.5--11.4\,m       & 400,695 \\
SSL4EO-L           & General Scenes   & Spectral                           & 30\,m              & 5 million \\
Satlas             & OD, CD, Class, SemSeg & RGB, spectral       & 1--10\,m           & 856K \\
GeoPile (GFM)      & General Scenes   & RGB                                          & 0.1--30\,m         & 587,954 \\
DOFA-data         & General Scenes   & RGB, SAR, spectral & 0.5--30\,m         & 8,081,411 \\
Prithvi            & General Scenes   & Spectral                & 30\,m              & 4.2 million \\
Skysense           & General Scenes   & RGB, SAR, Spectral                      & 0.31--20\,m        & 21.5 million \\
MMEarth            & General Scenes   & RGB, spectral, SAR, DEM                 & 10\,m              & 1.2 million \\
\bottomrule
\end{tabular}
\end{table*}

\subsubsection{\textbf{Foundation Models in RS Enable Transforming EO}}
Foundation models in RS are transformative tools designed to address the unique challenges associated with EO data. RS data is inherently complex, encompassing diverse modalities such as optical RGB, synthetic aperture radar (SAR), spectral data, LiDAR, and spatiotemporal sequences. Traditional RS models often struggle with integrating these multimodal datasets and require extensive labeled data for each specific application. Foundation models overcome these limitations by leveraging large-scale pretraining on diverse RS datasets, enabling the models to learn universal representations that are applicable across a broad spectrum of geospatial tasks. {Table \ref{tab:rs_datasets} summarizes widely used datasets for pretraining RS foundation models, emphasizing key characteristics such as domain focus, sensing modality, spatial resolution, data scale, and downstream suitability. Specifically, \textit{fMoW} and \textit{MillionAID} provide rich categorical annotations and medium-to-high resolution RGB imagery with limited spectral data coverage (0.3-11.4 m), making them particularly suitable for image- or scene-level classification and object-presence recognition. {\textit{SSL4EO-S12} (Sentinel-1/2, 10-60 m) \cite{wang2023ssl4eo} and \textit{SSL4EO-L} (Landsat, 30 m) \cite{stewart2023ssl4eo}} combine broad spectral coverage with regular revisit cycles, supporting large-scale semantic segmentation and land-cover mapping where moderate resolution is sufficient. \textit{Satlas} is designed for object detection, change detection, classification, and semantic segmentation, leveraging NAIP/RGB/Sentinel imagery (1-10 m) to address tasks requiring fine spatial detail and cross-source consistency. Broad ``general-scene'' corpora such as \textit{SeCo}, \textit{GeoPile (GFM)}, \textit{Prithvi}, \textit{DOFA-data}, \textit{Skysense}, and \textit{MMEarth} span diverse modalities (RGB, SAR, spectral, DEM) and resolutions (0.1-30 m), promoting robust transferability across downstream tasks. In particular, \textit{DOFA-data}, \textit{Skysense}, and \textit{MMEarth} incorporate SAR and/or DEM layers, offering advantages for applications that demand weather-robust sensing (SAR) or topographic context (DEM).}

These models utilize self-supervised or unsupervised learning techniques during the pretraining phase, allowing them to extract meaningful patterns and features from vast amounts of unlabeled RS data. This approach significantly reduces the dependency on labeled datasets, which are often limited in RS due to the challenges of pixel-level annotation across large and complex geographical areas. By enabling efficient transfer learning, foundation models can be fine-tuned with minimal labeled data to perform specific tasks, such as land cover classification, climate monitoring, disaster response, and agricultural assessment.

The adaptability of foundation models in RS allows them to handle varying data resolutions, sensor types, and temporal frequencies, making them suitable for dynamic and heterogeneous EO applications. Their scalability ensures that they can process large-scale datasets, such as continuous satellite imagery streams or global climate data, providing robust and timely insights for scientific research and practical applications.

\begin{figure*}[!t]
   \centering
		\includegraphics[width=1.0\textwidth]{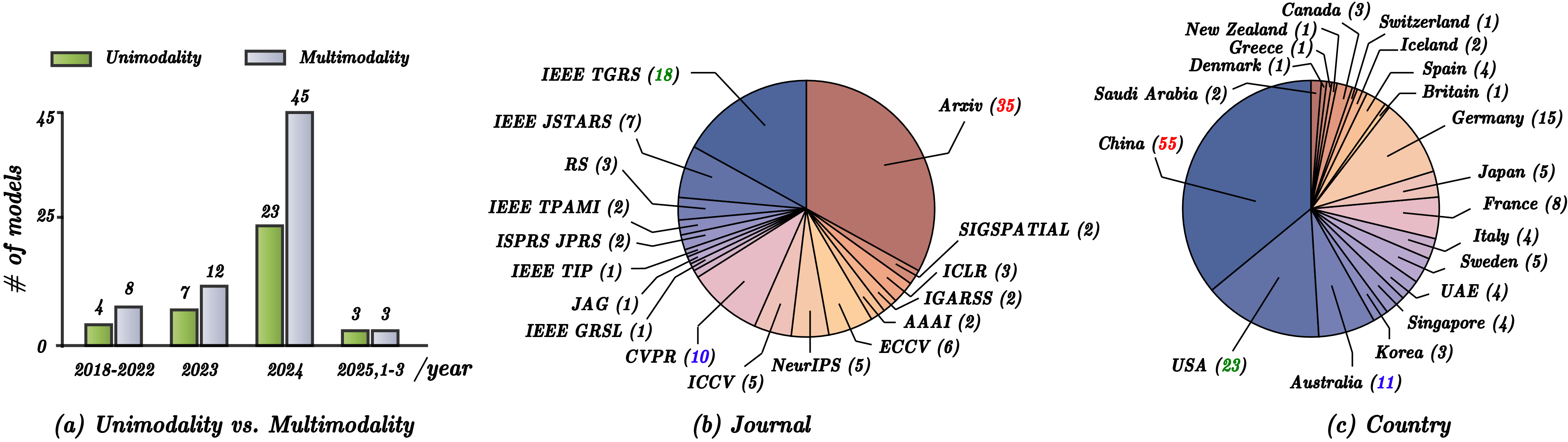}
    \caption{A statistical overview of foundation models in RS-related fields. (a) presents a quantitative comparison of unimodal and multimodal foundation models in RS. (b) illustrates the distribution of RS foundation models across different journals using a pie chart. (c) depicts the number of RS foundation models developed by different countries. The top three quantities in the two pie charts are highlighted in red, green, and blue.}
\label{fig:statistics}
\end{figure*}

\subsection{Learning Paradigm in RS}
\subsubsection{\textbf{Bottlenecks in Traditional RS Methods}}
The traditional learning-based paradigm in RS image processing and analysis typically involves two primary approaches. The first focuses on manually or automatically extracting feature representations from RS images, followed by the application of classification or regression algorithms to interpret these features for specific tasks such as land cover classification, object detection, or environmental monitoring. The second approach leverages end-to-end models, which directly map input data to output labels, streamlining the learning process by bypassing explicit feature extraction. Despite their effectiveness in certain applications, these traditional approaches are fundamentally limited by the availability of labeled data. In RS, labeling is particularly challenging due to the need to annotate vast amounts of data at the pixel level, often from a bird's-eye perspective that covers extensive geographic areas. This labor-intensive and time-consuming process results in a scarcity of large-scale labeled datasets, leaving a significant proportion of RS data unlabeled and underutilized.

\subsubsection{\textbf{Revolutionizing RS with Foundation Models}}
In contrast, the advent of foundation models introduces a transformative learning paradigm characterized by \textbf{\textit{``self-supervised pretraining + fine-tuning''}}. This approach has increasingly become the dominant framework in RS, offering a solution to the limitations posed by traditional methods. Self-supervised pretraining enables models to learn rich, generalized representations from massive amounts of unlabeled RS data by identifying inherent patterns, structures, and relationships within the data itself. This process leverages the natural redundancy and complexity of RS datasets, such as temporal sequences, multimodal inputs (e.g., optical, SAR, LiDAR), and spectral variations, to build robust feature representations without the need for manual annotations.

Following the pretraining phase, fine-tuning is employed to adapt the pretrained model to specific downstream tasks using a relatively small amount of labeled data. This significantly reduces the dependency on extensively labeled datasets while maintaining high performance across various applications. The fine-tuning process refines the model's representations to align with task-specific objectives, such as land cover classification, change detection, or object recognition, thereby enhancing accuracy and generalization capabilities.

The foundation model-based paradigm is particularly advantageous in RS scenarios due to the vast availability of unlabeled data from diverse sensors and platforms. By effectively utilizing this data, foundation models can overcome the data scarcity issue inherent in traditional methods, leading to more efficient and scalable solutions. Furthermore, the ability to generalize across different RS tasks and datasets makes foundation models highly versatile, supporting a wide range of applications from environmental monitoring to urban planning.

Fig. \ref{fig:paradigm} illustrates the general learning paradigm of foundation models in RS, highlighting the transition from traditional supervised approaches to the more flexible and powerful self-supervised pretraining and fine-tuning framework. This shift not only enhances the efficiency of RS data utilization but also paves the way for more advanced, generalized, and robust RS applications.

\begin{figure*}[!t]
   \centering
		\includegraphics[width=1.0\textwidth]{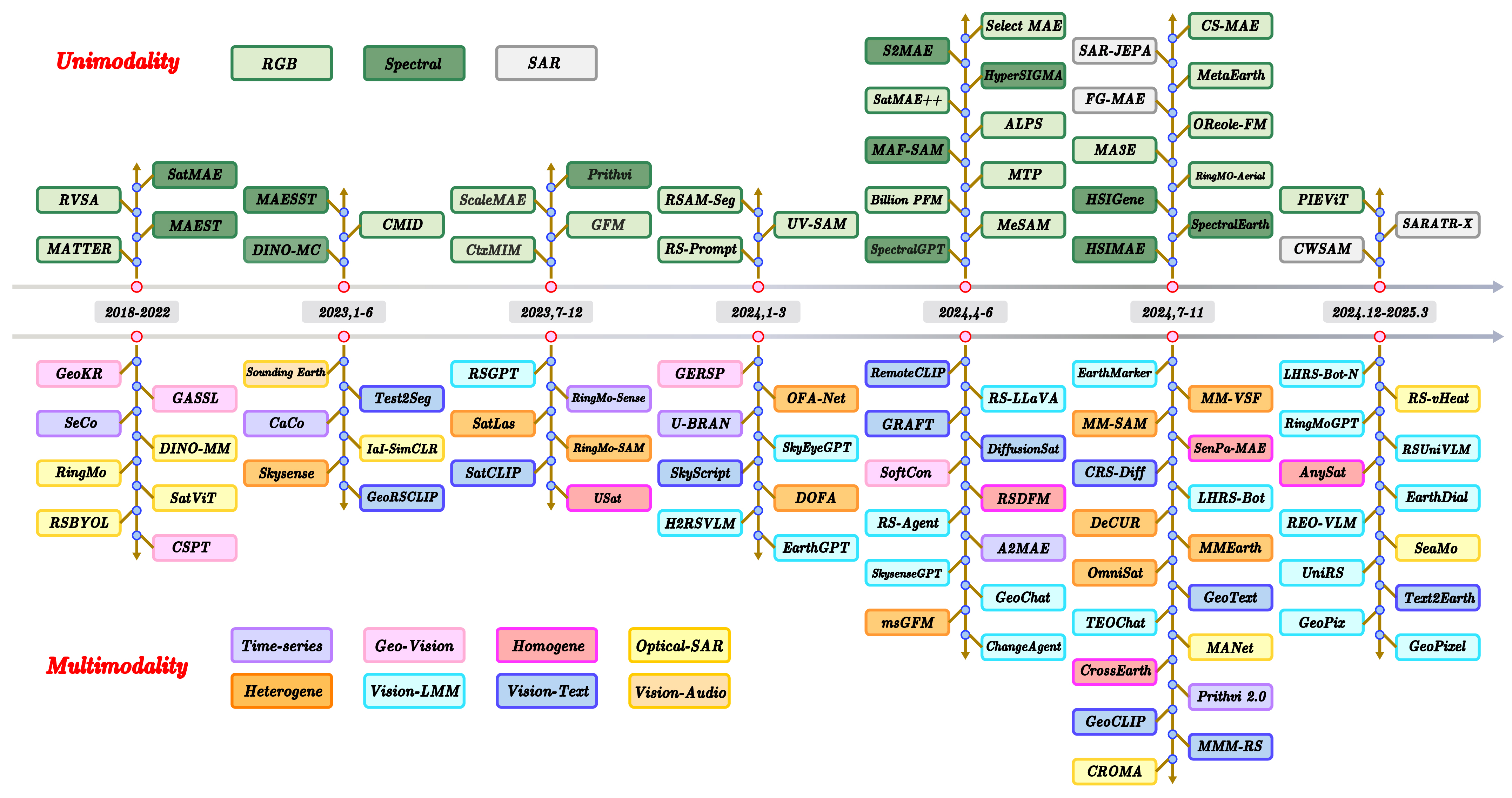}
    \caption{Roadmap of Foundation Models in RS from 2018 to 2025: The evolution from unimodal to multimodal approaches in RS. Unimodal data includes RGB, spectral, and SAR data. Multimodal integration is categorized into homogeneous data (e.g., multi-optical), optical-SAR, and heterogeneous data (e.g., combinations of optical-SAR-spectral-DSM, vision-text, vision-audio, geospatial-vision, time-series, and vision-LMM). Different model categories are color-coded to distinguish various data modalities and integration types, with models arranged in chronological order within each time period (vertical orientation).}
\label{fig:timeline}
\end{figure*}

\section{The evolution from unimodality to multimodality}
This section provides an intuitive overview of foundation models in RS, tracing their evolution from unimodality to multimodality. Through this perspective, we explore how these models increasingly integrate diverse data types, i.e., both unimodal and multimodal data. Fig. \ref{fig:timeline} illustrates the roadmap of foundation models in RS from unimodality to multimodality between 2018 and 2025. As shown in the figure, multimodal foundation models have gained increasing attention and popularity compared to unimodal models, particularly after 2024, which marked a significant surge in their development. Notably, vision foundation models integrated with large language models (LMMs) have garnered growing interest from researchers.

\subsection{Overview: Statistical Analysis}
We collected 105 papers on foundation models in RS-related fields and conducted a statistical analysis to better understand their trends and impact. Fig. \ref{fig:statistics} presents an overview of these models from three perspectives: modality distribution, publication venues, and country-wise contributions.

Fig. \ref{fig:statistics}(a) compares unimodal and multimodal RS foundation models, showing a clear shift toward multimodal approaches. This trend became particularly evident in 2024, as the rapid development of RS foundation models led to increased integration of diverse data sources, such as optical imagery, SAR, and spectral data. The growing interest in multimodal models reflects the need for more comprehensive and robust solutions in remote sensing applications.

Regarding publication distribution (see Fig. \ref{fig:statistics}(b)), arXiv hosts the largest number of papers, highlighting the early-stage dissemination of research findings before formal peer review. Among peer-reviewed venues, IEEE TGRS ranks second, emphasizing the strong influence of the geoscience and RS community, while CVPR, a premier AI and computer vision conference, takes third place. This indicates the increasing intersection of RS with cutting-edge AI and vision research.

Fig. \ref{fig:statistics}(c) shows the geographic distribution of the RS foundation model research. China leads in contributions, followed by the USA and Australia, demonstrating strong engagement from both traditionally dominant AI research hubs and emerging leaders in RS-focused AI. This distribution underscores the global nature of the RS foundation model development, driven by the need for advanced AI solutions in Earth observation and geospatial analysis.

Overall, these statistics highlight the rapid progress of foundation models in RS, the shift toward multimodality, and the increasing involvement of AI-driven methodologies. This trend reflects a growing commitment from the research community to push the boundaries of RS intelligence and develop more generalizable, scalable, and high-performance models for diverse applications.

\subsection{Unimodal Foundation Models in RS}
Unimodal foundation models in RS are designed to pretrain on data from a single source or sensor type, leveraging the unique characteristics of a specific data modality. Examples include high-resolution RGB imagery, radar data, point clouds, spectral data\footnote{We collectively refer to multispectral and hyperspectral data in RS as spectral data.}, and meteorological data. These models aim to extract the most informative features and patterns from their respective modalities, optimizing their capabilities to address distinct properties and challenges. This targeted approach lays a cornerstone for advancing RS tasks within specialized data domains.

\subsubsection{\textbf{Optical RGB Images}}
High-resolution spatial and semantic information in optical RGB images can be well represented via self-supervised and contrastive learning methods \cite{dou2024cc2vec}. For instance, MATTER \cite{akiva2022self} employs contrastive learning to emphasize texture information in RS images, while RVSA \cite{wang2022advancing} adopts a masking strategy inspired by MAE for pretraining on the MillionAID \cite{long2021creating} dataset, followed by fine-tuning using its RVSA module. CMID \cite{muhtar2023cmid} introduces a contrastive mask image distillation method by combining contrastive learning (CL) with mask image modeling (MIM) for global semantic separability and local spatial perceptibility. CtxMIM \cite{zhang2023ctxmim} introduces a context-enhanced generative branch to provide contextual information via consistency constraints for patch-wise reconstruction. The GFM model \cite{mendieta2023towards} leverages a large-scale public dataset and a teacher-student dual-stream network with masked image reconstruction as its objective, achieving notable success in scene classification, change detection, and semantic segmentation tasks.

Models such as Scale-MAE \cite{reed2023scale} and SatMAE++ \cite{noman2024rethinking} both address image size and resolution in RS: Scale-MAE focuses on scale-invariant representation learning, while SatMAE++ enhances resolution-aware feature extraction. Consequently, Scale-MAE benefits cross-scale generalization tasks, whereas SatMAE++ is advantageous for high-resolution applications such as object detection and fine-grained segmentation. SelectiveMAE \cite{wang2024scaling} dynamically encodes and reconstructs image patches selected according to semantic richness, employing a progressive semantic token selection module that evolves from reconstructing semantically similar tokens to encoding complementary semantic dependencies. MA3E \cite{li2024masked} proposes a mask self-supervised learning framework based on rotational angle variations, addressing the differences between RS images and natural images. RingMo-Aerial \cite{diao2024ringmo} incorporates spectral information as an auxiliary, training through contrastive learning. Billion PFM \cite{cha2024billion}, and OReole-FM \cite{dias2024oreole} extend the pretraining of high-resolution RS images to billion-parameter models using the MAE approach. The work in \cite{yu2024metaearth} proposes a resolution-guided self-cascading generative framework, called MetaEarth, which is capable of generating images of arbitrary regions with a wide range of geographic resolutions. Cross-Scale MAE \cite{tang2023cross} performs contrastive mask modeling on images of different sizes, forming a unified optimization framework for pretraining.  PIEViT \cite{lu2025pattern} is a novel self-supervised learning framework designed for RS images, employing a teacher-student architecture to handle both image-level and block-level tasks.

Additionally, the segmentation capabilities of the segment anything model (SAM) have garnered widespread attention in the RS field. MeSAM \cite{zhou2024mesam} designs a new adapter for RS optical images, embedding it within the SAM encoder to process images more effectively using multi-scale convolutional kernels while preserving high-frequency features. RSAM-Seg \cite{zhang2024rsam} enhances SAM by incorporating domain knowledge and a scaling module, performing tasks such as cloud, building, field, and road segmentation. Rsprompter \cite{chen2024rsprompter} introduces a learnable anchor-based prompt and mask decoder, enabling SAM to handle class inputs for instance segmentation. UV-SAM \cite{zhang2024uv} refines the coarse masks generated by class activation maps into pseudo-labels for training semantic segmentation models on RS images. ALPS \cite{zhang2024alps} proposes a framework for fully automatic dense annotation using SAM without the need for prior knowledge or prompts.

\subsubsection{\textbf{Spectral Data}}
Various models leverage multi-temporal and multi-band data to improve spatial and spectral information reconstruction \cite{hong2026hyperspectral}. SatMAE \cite{cong2022satmae} introduces a grouping mask strategy tailored for multi-temporal, multichannel data, enabling better learning of channel and spatial relationships. DINO-MC \cite{wanyan2024extending} employs a contrastive learning framework, and Prithvi \cite{hsu2024geospatial} enhances spectral-spatial representation learning by masking individual channels. MAF-SAM \cite{song2024multispectral} proposes a multi-stage adaptive fine-tuning strategy, compressing spectral data into three channels using prefix adapters, applying LoRA fine-tuning to inject crop-specific knowledge into the image encoder, and generating multi-class segmentation masks with precise category information. Beyond previous approaches, SpectralGPT \cite{hong2024spectralgpt} is the first foundation model specifically tailored for spectral RS data. By explicitly modeling the intrinsic spatial-spectral coupling and spectral sequentiality of multi/hyper-spectral observations, SpectralGPT introduces a 3D masking strategy together with spectral consistency constraints. This design enables effective learning of spectral sequences and structured spectral representations, thereby substantially advancing both the modeling paradigm and practical understanding of spectral remote sensing data. MAEST \cite{ibanez2022masked} and MAESST \cite{scheibenreif2023masked} introduce masked self-supervised learning algorithms tailored to spectral characteristics to enhance land cover classification accuracy. Wang et al. \cite{wang2024hsimae} performed self-supervised mask modeling pre-training on the large-scale spectral cube. The SpectralEarth \cite{braham2024spectralearth} introduces a self-supervised learning algorithm, yielding a solid data foundation and algorithmic support for pretraining EnMap data with a series of fine-tuning downstream applications. HyperSIGMA \cite{wang2024hypersigma} introduced a two-stream framework that separately models spatial and spectral information. In this approach, input spectral data are unified to a fixed number of bands by randomly selecting spectral bands and shuffling the spectral sequence. Pang et al. \cite{pang2024hsigene} proposed a generation foundation model, which is based on a latent diffusion model and supports multi-condition control. It also designs a two-stage super-resolution framework, generating the realism of the generated images.

\subsubsection{\textbf{Synthetic Aperture Radar (SAR)}}
FG-MAE \cite{wang2024feature} leverages histograms of oriented gradients (HOGs) as reconstruction features, utilizing the spatial information inherent in SAR data. SAR-JEPA \cite{li2024predicting} adopts a joint embedding prediction architecture to predict gradient information for pre-training on SAR images. CWSAM \cite{pu2024classwise} adapts the SAM framework for land cover classification in SAR images by incorporating task-specific input modules and a categorized mask decoder while injecting SAR low-frequency information for the segmentation task. SARATR-X \cite{li2025saratr} employs a multi-scale masking strategy for pre-training on SAR images, achieving object detection and classification tasks.

\subsection{Multimodal Foundation Models in RS}
While unimodal models in RS typically focus on single data types, such as optical, radar, or LiDAR, integrating multimodal data introduces an additional layer of complexity due to the heterogeneous nature of RS data. These data often come from different sensors, platforms, and scales, leading to variations in spatial, temporal, and spectral characteristics. For example, optical, radar, and LiDAR data each offer distinct perspectives of the Earth's surface, but their integration requires addressing differences in resolution, data format, and noise characteristics. Moreover, combining both structured data (such as time series or geospatial coordinates) and unstructured data (such as text or images) further complicates the process. One of the main challenges of multimodal integration lies in harmonizing these diverse data types into a unified model that accurately captures the underlying phenomena while preserving the strengths of each modality. Furthermore, this integration must account for the varying degrees of uncertainty and incompleteness inherent in each data type, making it more challenging to develop robust and reliable models.

Multimodal foundation models in RS extend beyond single modalities by integrating complementary information from diverse data types. These models handle homogeneous and heterogeneous data and combine visual data with other modalities, such as time series, geospatial, text, LMMs, and audio. Unlike multi-source data, which emphasizes different sources of RS data, multimodality focuses more on the integration of diverse data types or forms, enabling a more comprehensive understanding of EO data. This approach enhances model performance by capturing the unique characteristics and relationships within each modality, ultimately improving the accuracy and robustness of RS applications. When working with multimodal RS data, it is essential to distinguish between homogeneous and heterogeneous sources. Homogeneous data (from the same sources but different sensors, such as Sentinel-2, Landsat, EnMap, GF-1/2/5/6) can typically be fused directly at the data level using operations such as band stacking, resampling, or resolution enhancement \cite{vivone2024deep}. In contrast, heterogeneous data (e.g., optical and SAR, or optical and DEM) generally cannot be combined at the raw data level due to differences in sensing principles, resolutions, and noise characteristics. In such cases, feature-level fusion is preferred, including early feature concatenation, intermediate feature interaction, and late decision-level combination. More recently, advanced interactive fusion strategies, such as cross-modal attention, graph-based reasoning, and co-learning mechanisms, have been introduced to dynamically capture cross-modal complementarities and thereby enhance performance on complex downstream tasks. For a more detailed discussion of best practices, readers may refer to \cite{hong2020more}.

\subsubsection{\textbf{Heterogeneous Data $\mapsto$ Time-Series}}
In the fusion of visual and temporal data, several models have introduced novel strategies to enhance spatiotemporal representation in RS. SeCo \cite{manas2021seasonal} explicitly accounts for the seasonal characteristics of RS data by constructing positive and negative sample pairs from different seasonal images, improving the model’s sensitivity to seasonal variations. However, SeCo struggles with recognizing long-term temporal invariances. To address this, CACo \cite{mall2023change} enhances temporal consistency with a contrastive loss function to extract more robust invariant features from time-series data.

Further advancements extend to RS video data. RingMo-Sense \cite{yao2023ringmo} targets aerial videos by employing multi-strategy masking for pretraining. Meanwhile, U-BARN \cite{dumeur2024self} adopts a BERT-like generative pretraining task, where masked portions of optical satellite image sequences are reconstructed.

Other models refine spatiotemporal encoding through advanced masking strategies. A$^2$-MAE \cite{zhang20242} integrates an anchor-aware masking strategy with a geo-encoding module, effectively capturing spatial, temporal, and spectral variations. Finally, Prithvi-EO-2.0 \cite{szwarcman2024prithvi} leverages time-series data from Landsat and Sentinel-2 via channel-masking pretraining to extract temporal features from homogeneous datasets, further strengthening time-series modeling in RS.

\subsubsection{\textbf{Heterogeneous Data $\mapsto$ Geospatial-Vision}}
GeoKR \cite{li2021geographical} utilizes cartographic information as labels, combining distillation and contrastive learning for semi-supervised pretraining. GASSL \cite{ayush2021geography} applies contrastive learning to uniformly augmented image data, incorporating geographic location as auxiliary supervision. CSPT \cite{zhang2022consecutive} proposes a contrastive spatial pretraining approach, employing dual encoders to encode images and geographic locations separately. Through contrastive objectives, it extracts effective location information from images for downstream supervised tasks such as image classification. Mai \textit{et al.} \cite{mai2023csp} proposed the CSP model by introducing a self-supervised contrastive learning framework for geospatial-visual representations. GERSP \cite{huang2024generic} employs a teacher-student distillation network, aligning natural and RS images geographically to enhance the perception of general land cover knowledge in pretraining models. SoftCon \cite{wang2024multi} introduces soft contrastive learning, leveraging multi-label supervision signals derived from land cover data to optimize cross-scene soft similarity metrics, enhancing the model's generalization in complex RS images.

\subsubsection{\textbf{Homogeneous Data}}
For homogeneous data, the USat \cite{irvin2023usat} model features a unique block projection layer and position encoding, enabling it to handle spectral data from multiple sensors and adapt to different spatial scales. RSDFM \cite{wang2024rs} employs the BEVFormer architecture to build a unified perspective from data across various platforms, angles, and altitudes. SenPa-MAE \cite{prexl2024senpa} uses spectral wavelength information as a prior embedding for the network and applies mask modeling self-supervised learning, allowing the model to perform exceptionally well on data from different spectral sensors. CrossEarth \cite{gong2024crossearth} introduces a domain generalization method to perform the semantic segmentation of foundation models across downstream datasets with different styles. AnySat \cite{astruc2024anysat} is a multimodal model that explores a resolution-adaptive spatial encoder. With the use of a scale-adaptive patch encoding module, AnySat can improve the model's sensitivity to resolution effectively, specifically for optical RS data with varying resolutions. CSMAE \cite{hackstein2024exploring} trained a cross-sensor MAE model for multisensor RS image retrieval to search semantically similar images across different image modalities effectively.

\subsubsection{\textbf{Heterogeneous Data $\mapsto$ Optical-SAR}}
RingMo \cite{sun2022ringmo} adopts a sparse masking strategy and Transformer architecture tailored to the characteristics of heterogeneous RS images. These models, e.g., IaI-SimCLR \cite{prexl2023multi}, DINO-MM \cite{wang2022self}, RS-BYOL \cite{jain2022self}, employ the contrastive learning paradigm in the context of multimodal, multi-source satellite imagery. SatViT \cite{fuller2022satvit} directly transfers the MAE framework to learn heterogeneous data from optical and SAR sources. RS-BYOL \cite{jain2022self} learns invariant feature embeddings from spectral and SAR data by combining single-channel and three-channel feature extraction. MANet \cite{ma2024manet} introduces two adapters into the SAM encoder for extracting single-modal information and the other for fusing multimodal information. CROMA \cite{fuller2024croma} utilizes the CoCa framework, employing contrastive masked pretraining to combine spectral and radar data, thus improving the model’s ability to handle data heterogeneity. RS-vHeat \cite{hu2024rs} explores using a heat conduction-based parallel computing model to simulate local region correlations in high-resolution RS images. Recently, Li et al. \cite{li2026seamo} proposed a novel multimodal foundation model in RS, called SeaMo, by embedding seasonal knowledge to improve the performance on a wide range of downstream applications.

\subsubsection{\textbf{Heterogeneous Data $\mapsto$ Optical/SAR/Spectral/DSM}}
In the realm of multimodal RS, several models have been developed to integrate diverse data sources effectively. SkySense \cite{guo2024skysense} employs a decomposed multimodal spatiotemporal encoder to handle optical and SAR time series data, leveraging multi-granularity contrastive learning to enhance its understanding of complex RS datasets. Similarly, SatLas \cite{bastani2023satlaspretrain} utilizes the Swin Transformer for large-scale supervised pretraining on NAIP, LiDAR, and Sentinel-2 data, improving its capacity to process diverse geospatial inputs.

For segmentation tasks, RingMo-SAM \cite{yan2023ringmo} enhances multimodal segmentation by freezing the SAM encoder and applying parameter-efficient fine-tuning (PEFT) with LoRA, while incorporating modality-specific decoders for RGB, SAR, PolSAR, DSM, and MSI data. Expanding on multimodal tokenization, OFA-Net \cite{xiong2024one} introduces dedicated tokenizers for different input types (RGB, spectral, and radar), ensuring the model’s sensitivity to diverse spectral and spatial information. Meanwhile, DOFA \cite{xiong2024neural} employs a dynamic weight generator to adaptively embed features across various spectral bands for creating an efficient foundation model that supports flexible input from RGB, spectral, and radar data.

To further enhance cross-sensor learning, msGFM \cite{han2024bridging} adopts a cross-sensor pretraining approach based on masked image modeling, enabling joint representations from optical, SAR, spectral, and DSM data, thus improving multimodal integration. Similarly, MM-VSF \cite{ravirathinam2024towards} combines spectral imagery and weather data within a knowledge-guided framework, which uses variable step forecasting as its pretraining objective to boost environmental prediction capabilities.

For multimodal segmentation, MM-SAM \cite{xiao2024segment} extends SAM to support cross-modal and multimodal processing, significantly improving segmentation performance in multisensor RS applications. Meanwhile, DeCUR \cite{wang2024decoupling} introduces a decoupling strategy to separate common and unique representations to effectively fuse complementary information across modalities, such as radar-optical, RGB-elevation, and RGB-depth.

Several models focus on holistic multimodal fusion and data assimilation. MMEarth \cite{nedungadi2024mmearth} reconstructs $\ell_{2}$ distinct pixel-level and region-level modalities and labels during pretraining, providing a comprehensive perspective for downstream tasks. Finally, OmniSat \cite{astruc2025omnisat} integrates high-resolution RGB, spectral time series, and radar time series data by leveraging heterogeneous modules with contrastive masking strategies for pretraining, achieving strong performance across a wide range of multimodal applications.

\subsubsection{\textbf{Heterogeneous Data $\mapsto$ Vision-LMM}}
The rapid development of LMMs has expanded RS tasks beyond traditional visual models, enabling more sophisticated multimodal interactions. A common approach involves leveraging visual-language foundation models by aligning visual encoders with textual semantics to facilitate integration with LLMs. For instance, RSGPT \cite{hu2025rsgpt} utilizes the EVA-CLIP encoder and is trained with advanced techniques in RS tasks. RS-LLaVA \cite{bazi2024rs} applies LoRA fine-tuning to significantly boost LLaVA's performance. EarthGPT \cite{zhang2024earthgpt} constructs a vision-enhanced perception mechanism that optimizes the fusion of coarse-scale semantic understanding with fine-scale detail extraction.

Beyond static image analysis, interactive and conversational models play a crucial role in RS applications. RS-Agent \cite{xu2024rs} positions the LLM as a ``central controller'', integrating multiple high-performance RS image-processing tools to enable multi-tool, multi-turn interactions. GeoChat \cite{kuckreja2024geochat} pioneers the first conversational VLGFM supporting visual geolocation tasks, introducing a new paradigm for interactive RS image analysis. Change-Agent \cite{liu2024change} proposes an interactive agent capable of performing comprehensive change detection, object counting, and causal analysis based on user instructions. EarthMarker \cite{zhang2024earthmarker} designs a cross-domain staged learning strategy to achieve multi-granular visual perception for image-level, region-level, and point-level RS interpretations.

Several models focus on RS video analysis and temporal reasoning. SkyEyeGPT \cite{zhan2025skyeyegpt} is the VLGFM designed for RS video captioning, offering novel solutions for automated video content description. TEOChat \cite{irvinteochat} emerges as a visual language model capable of interacting with time-series RS data, excelling in various temporal reasoning tasks. UniRS \cite{li2024unirs} is the unified visual-language model for multi-temporal RS tasks, supporting single images, image pairs, and videos for comprehensive temporal analysis in a unified framework.

To refine RS image comprehension, LHRS-Bot \cite{muhtar2024lhrs} employs hierarchical visual-language alignment strategies and curriculum learning methods to provide new perspectives in RS analysis. LHRS-Bot-Nova \cite{li2024lhrs} enhances LHRS-Bot with a vision encoder strengthened by a mixture of expert models and a new bridging module, improving multimodal large language model comprehension of RS images for interactive natural language dialogue. RingMoGPT \cite{wang2024ringmogpt} addresses key issues in RS, such as object-level recognition, localization, and multi-temporal change detection, with a location- and instruction-aware Q-Former and change detection module for target detection and change description. RSUniVLM \cite{liu2024rsunivlm} introduces a Granularity-oriented Mixture of Experts, enhancing the model’s ability to capture visual information at different granularities.

Emerging models emphasize high-resolution and fine-grained perception. GeoPix \cite{ou2025geopix} extends the multimodal large language model understanding of RS images to the pixel level. GeoPixel \cite{shabbir2025geopixel} is an end-to-end high-resolution multimodal model that supports pixel-level localization, enabling fine-grained visual perception through interleaved masks in dialogue. It supports high-resolution RS image analysis up to 4K and is compatible with any aspect ratio. EarthDial \cite{soni2024earthdial} supports spectral, multi-temporal, and multi-resolution images, performing a wide range of RS tasks such as classification, detection, description generation, question answering, visual reasoning, and visual localization.

Some models integrate domain-specific AI with Earth observation. REO-VLM \cite{xue2024reo} combines regression capabilities with traditional generative functions, using language-driven reasoning to incorporate scientific domain knowledge, enabling complex interpretation of Earth observation data and supporting tasks like biomass regression and image content explanation. SkySenseGPT \cite{luo2024skysensegpt} incorporates principles from computer graphics to enhance the model’s ability to perceive and reason relationships between RS targets. These developments represent a shift toward more interactive, domain-aware, and high-resolution AI-driven RS analysis, marking a new era in geospatial intelligence.

\begin{table*}
\centering
\caption{A concise comparison of existing unimodal foundation models (e.g., RGB, spectral data, SAR) in RS, detailing their corresponding publications, pretrained architectures, downstream applications, and code availability. CL: Contrastive Learning, MAE: Msked Autoencoders, MIM: Masked Image Modeling, Diff: Diffusion Model, JEPA: Joint Embedding Predictive Architecture, SC: Scene Classification, SS: Semantic Segmentation, CD: Change Detection, OD: Object Detection, SR: Super-resolution, AD: Anomaly Detection, SU: Spectral Unmixing, DN: Denoising, IG: Image Generation.}
\resizebox{\textwidth}{!}{
\begin{tabular}{c|c|c|c|c|c}
\toprule[1.5pt]
Modality & Models & Publications & Architectures & Downstreams & Codes\\
\hline\hline
\addlinespace[2pt]
\multirow{16}{*}{RGB} & \href{https://openaccess.thecvf.com/content/CVPR2022/html/Akiva_Self-Supervised_Material_and_Texture_Representation_Learning_for_Remote_Sensing_Tasks_CVPR_2022_paper.html}{Matter}~\cite{akiva2022self} & CVPR'23 & CL & SC, SS, CD & \href{https://github.com/periakiva/MATTER}{https://github.com/periakiva/MATTER}\\
& \href{https://openaccess.thecvf.com/content/ICCV2023/papers/Reed_Scale-MAE_A_Scale-Aware_Masked_Autoencoder_for_Multiscale_Geospatial_Representation_Learning_ICCV_2023_paper.pdf}{Scale-MAE}~\cite{reed2023scale} & ICCV'23 & MAE & SC, SS & \href{https://github.com/bair-climate-initiative/scale-mae}{https://github.com/bair-climate-initiative/scale-mae} \\
& \href{https://ieeexplore.ieee.org/abstract/document/9956816}{RVSA}~\cite{wang2022advancing} & TGRS'22 & MIM & SC, SS, OD & \href{https://github.com/ViTAE-Transformer/Remote-Sensing-RVSA}{https://github.com/ViTAE-Transformer/Remote-Sensing-RVSA}\\
& \href{https://ieeexplore.ieee.org/abstract/document/10105625}{CMID}~\cite{muhtar2023cmid} & TGRS'23 & CL\&MIM & SC, CD, OD & \href{https://github.com/NJU-LHRS/official-CMID}{https://github.com/NJU-LHRS/official-CMID}\\
& \href{https://openaccess.thecvf.com/content/ICCV2023/papers/Mendieta_Towards_Geospatial_Foundation_Models_via_Continual_Pretraining_ICCV_2023_paper.pdf}{GFM}~\cite{mendieta2023towards} & ICCV'23 & MIM & SC, SS, CD, SR & \href{https://github.com/mmendiet/GFM}{https://github.com/mmendiet/GFM} \\
& \href{https://openreview.net/pdf?id=5oEVdOd6TV}{Cross-Scale MAE}~\cite{tang2023cross} & NeurIPS'23 & MAE & SC, SS & \href{https://github.com/aicip/Cross-Scale-MAE}{https://github.com/aicip/Cross-Scale-MAE} \\
& \href{https://ieeexplore.ieee.org/document/10547536}{MTP}~\cite{wang2024mtp} & JSTARS'24 & Supervised & SS, CD, OD & \href{https://github.com/ViTAE-Transformer/MTP}{https://github.com/ViTAE-Transformer/MTP}\\
& \href{https://arxiv.org/abs/2406.11933}{Selective MAE}~\cite{wang2024scaling} & Arxiv'24 & MAE & SC, SS, OD & \href{https://github.com/Fengxiang23/SelectiveMAE}{https://github.com/Fengxiang23/SelectiveMAE}\\
& \href{https://www.ecva.net/papers/eccv_2024/papers_ECCV/papers/01288.pdf}{MA3E}~\cite{li2024masked} & ECCV'24 & MAE & SC, SS, OD & \href{https://github.com/benesakitam/MA3E}{https://github.com/benesakitam/MA3E}\\
& \href{https://openaccess.thecvf.com/content/CVPR2024/papers/Noman_Rethinking_Transformers_Pre-training_for_Multi-Spectral_Satellite_Imagery_CVPR_2024_paper.pdf}{SatMAE++}~\cite{noman2024rethinking} & CVPR'24 & MAE & SC & \href{https://github.com/techmn/satmae_pp}{https://github.com/techmn/satmae\_pp}\\
& \href{https://ieeexplore.ieee.org/document/10409216}{RS-Prompter}~\cite{chen2024rsprompter} & TGRS'24 & Supervised & SS & \href{https://github.com/KyanChen/RSPrompter}{https://github.com/KyanChen/RSPrompter}\\
& \href{https://ojs.aaai.org/index.php/AAAI/article/view/30260}{UV-SAM}~\cite{zhang2024uv} & AAAI'24 & Supervised & SS & \href{https://github.com/tsinghua-fib-lab/UV-SAM}{https://github.com/tsinghua-fib-lab/UV-SAM}\\
& \href{https://www.mdpi.com/2072-4292/17/4/590}{RSAM-Seg}~\cite{zhang2024rsam} & RS'24 & Supervised & SS & \href{https://github.com/Chief-byte/RSAM-Seg}{https://github.com/Chief-byte/RSAM-Seg}\\
& \href{https://arxiv.org/abs/2406.10855}{ALPS}~\cite{zhang2024alps} & {TIP'24} & Supervised & SS & \href{https://github.com/StriveZs/ALPS}{https://github.com/StriveZs/ALPS} \\
& \href{https://ieeexplore.ieee.org/document/10522788}{MeSAM}~\cite{zhou2024mesam} & TGRS'24 & Supervised & SS & \href{https://github.com/Magic-lem/MeSAM}{https://github.com/Magic-lem/MeSAM}\\
& \href{https://ieeexplore.ieee.org/document/10768939}{MetaEarth}~\cite{yu2024metaearth}
& TPAMI'24 & Diff & IG & \href{https://jiupinjia.github.io/metaearth/}{https://jiupinjia.github.io/metaearth/} \\
\addlinespace[2pt]
\hline\hline
\addlinespace[2pt]
\multirow{8}{*}{Spectral Data} & \href{https://papers.nips.cc/paper_files/paper/2022/file/01c561df365429f33fcd7a7faa44c985-Paper-Conference.pdf}{SatMAE}~\cite{cong2022satmae} & NeurIPS'22 & MAE & SC,SS & \href{https://sustainlab-group.github.io/SatMAE/}{https://sustainlab-group.github.io/SatMAE/}\\
& \href{https://ieeexplore.ieee.org/document/10678231}{DINO-MC}~\cite{wanyan2024extending} & CVPRW'24 & CL & SC, CD & \href{https://github.com/WennyXY/DINO-MC}{https://github.com/WennyXY/DINO-MC}\\
& \href{https://ieeexplore.ieee.org/document/10490262}{SpectralGPT}~\cite{hong2024spectralgpt} & TPAMI'24 & MAE & SC, SS, CD & \href{https://github.com/danfenghong/IEEE_TPAMI_SpectralGPT}{https://github.com/danfenghong/IEEE\_TPAMI\_SpectralGPT} \\
& \href{https://ieeexplore.ieee.org/abstract/document/9931741}{MAEST}~\cite{ibanez2022masked} & TGRS'22 & MAE & SS & \href{https://github.com/ibanezfd/MAEST}{https://github.com/ibanezfd/MAEST} \\
& \href{https://ieeexplore.ieee.org/document/10208365}{MAESST}~\cite{scheibenreif2023masked} & CVPRW'23 & MAE & SS & \href{https://github.com/HSG-AIML/MaskedSST/}{https://github.com/HSG-AIML/MaskedSST/} \\
& \href{https://arxiv.org/abs/2406.11519}{HyperSIGMA}~\cite{wang2024hypersigma} & {TPAMI'25} & MAE & SS, AD, OD, CD, SU, SR, DN & \href{https://github.com/WHU-Sigma/HyperSIGMA}{https://github.com/WHU-Sigma/HyperSIGMA} \\
& \href{https://ieeexplore.ieee.org/stamp/stamp.jsp?arnumber=10607879}{HSIMAE}~\cite{wang2024hsimae} & JSTARS'24 & MAE & SS & \href{https://github.com/Ryan21wy/HSIMAE}{https://github.com/Ryan21wy/HSIMAE} \\
& \href{https://arxiv.org/abs/2409.12470}
{HSIGene}~\cite{pang2024hsigene} & Arxiv'24 & Diff & IG & \href{https://github.com/LiPang/HSIGene}{https://github.com/LiPang/HSIGene} \\
\addlinespace[2pt]
\hline\hline
\addlinespace[2pt]
\multirow{4}{*}{SAR} & \href{https://ieeexplore.ieee.org/stamp/stamp.jsp?arnumber=10766851}{FG-MAE}~\cite{wang2024feature} & JSTARS'24 &MAE & SC, SS & \href{https://github.com/zhu-xlab/FGMAE}{https://github.com/zhu-xlab/FGMAE} \\
& \href{https://www.sciencedirect.com/science/article/pii/S0924271624003514}{SAR-JEPA}~\cite{li2024predicting} & ISPRS'24 & JEPA & OD & \href{https://github.com/waterdisappear/SAR-JEPA}{https://github.com/waterdisappear/SAR-JEPA} \\
& \href{https://ieeexplore.ieee.org/stamp/stamp.jsp?arnumber=10849617}{CWSAM}~\cite{pu2024classwise} & JSTARS'25 & Supervised & SS & \href{https://github.com/xypu98/CWSAM}{https://github.com/xypu98/CWSAM} \\
& \href{https://ieeexplore.ieee.org/abstract/document/10856784}{SARATR-X}~\cite{li2025saratr} & TIP'25 & MAE & SC, OD & \href{https://github.com/waterdisappear/SARATR-X}{https://github.com/waterdisappear/SARATR-X} \\
\bottomrule[1.5pt]
\end{tabular}
}
\label{tab:unimodal}
\end{table*}

\subsubsection{\textbf{Heterogeneous Data $\mapsto$ Vision-Text}}
In the vision-text domain, several models have been developed to bridge the gap between RS images and textual descriptions. Text2Seg \cite{zhang2024text2seg} integrates the zero-shot segmentation capability of SAM with CLIP and Grounding DINO, employing a two-stage framework, e.g., PreSAM and PostSAM, to achieve text-guided semantic segmentation of RS images. Similarly, GeoRSCLIP \cite{zhang2024rs5m}, released alongside the RS5M dataset, is a vision-language-pretrained model specifically designed for RS imagery. Inspired by CLIP, {Li. et al. \cite{li2023rs} introduced a vision-language model (RS-CLIP) for RS scene classification based on contrastive vision-language supervision, while} SatCLIP \cite{klemmer2023satclip} enhances the alignment between images and their geographic locations, improving the model’s geospatial awareness.

Beyond direct image-text alignment, other models leverage external geospatial information to enrich RS image understanding. SkyScript \cite{wang2024skyscript} utilizes OpenStreetMap data to annotate RS images based on Ground Sampling Distance (GSD), enabling a more structured approach to geospatial labeling. Meanwhile, RemoteCLIP \cite{liu2024remoteclip} is a pretrained model for RS image-language tasks, constructing image-text pairs from existing datasets to enhance retrieval, detection, and segmentation capabilities. To address the scarcity of text-annotated RS data, GRAFT \cite{mall2023remote} pairs georeferenced ground-level images from Flickr with corresponding RS images and captions, effectively augmenting the available training data.

In the realm of generative modeling, diffusion-based approaches have emerged as powerful tools for RS applications. DiffusionSat \cite{khanna2024diffusionsat} integrates metadata, such as geographic location in the diffusion model, tackling tasks like time series generation, super-resolution reconstruction from spectral inputs, and image restoration on large-scale high-resolution datasets. Expanding on this, CRS-Diff \cite{tang2403crs} introduces a multi-condition control mechanism, allowing precise generation by supporting text, metadata, and image inputs. GeoText \cite{chu2024towards} further enhances spatial and fine-grained perception by aligning land cover targets with text descriptions, improving the interpretability of geospatial datasets.

Several models also explore novel strategies for image-location alignment and multimodal data fusion. GeoCLIP \cite{vivanco2023geoclip}, inspired by image-to-GPS retrieval strategies, aligns RS images with their corresponding GPS locations as a continuous function by leveraging positional encoding in the Fourier domain. MMM-RS \cite{luo2024mmm} advances text-to-image generation by training a CycleGAN-based model on large-scale multimodal, multi-resolution RS imagery. Finally, Text2Earth \cite{liu2025text2earth}, a 1.3-billion-parameter generative model built on a diffusion framework, is designed specifically for global-scale RS scene modeling, further pushing the boundaries of text-conditioned RS image synthesis.

\subsubsection{\textbf{Heterogeneous Data $\mapsto$ Vision-Audio}}
In vision-audio fusion, SoundingEarth \cite{heidler2023self} leverages audio data matched with RS images to construct training pairs. Within a contrastive learning framework, it aligns audio and image representations to enable effective pretraining for tasks requiring multimodal understanding.

\begin{table*}[!t]
\centering
\setlength{\tabcolsep}{1.5pt}
\renewcommand{\arraystretch}{1.15}
\caption{Performance comparison of unimodal and multimodal models on PANGAEA benchmark. The best-performing model for each dataset is highlighted in red, the second-best in blue, and the third-best is underlined.}
\label{tab:multitask_performance}
\begin{tabular}{llccccccccccc}
\toprule
\textbf{Modality} & \textbf{Model} & \textbf{BurnSr} & \textbf{MADOS} & \textbf{PASTIS} & \textbf{Sen1Fl11} & \textbf{FBP} & \textbf{DEN} & \textbf{CTM-SS} & \textbf{SN7} & \textbf{AI4Farms} & \textbf{Xview2} & \textbf{BioMassters} \\
& & \textit{Wildfire} & \textit{Marine} & \textit{Agriculture} & \textit{Flood} & \textit{Land} & \textit{Land} & \textit{Agriculture} & \textit{CD} & \textit{Agriculture} & \textit{CD} & \textit{Regression} \\
\midrule
\multirow{6}{*}{\textbf{Unimodal}} & SatlasNet   & 79.96 & 55.86 & 17.51 & 90.30 & 50.97 & 36.31 & 46.97 & {61.88} & 25.13 & 52.23 & \underline{41.67} \\
& Prithvi 1.0 & {83.62} & 49.98 & 33.93 & \underline{90.37} & 46.81 & 27.86 & 43.07 & 56.54 & \underline{26.86} & 49.35 & 39.99 \\
& SpectralGPT & 80.47 & 57.99 & \underline{35.44} & 89.07 & 33.42 & 37.85 & 46.95 & 58.86 & 26.75 & 48.40 & 36.11 \\
& S.-S12-MoCo & 81.58 & 51.76 & 34.49 & 89.26 & \underline{53.02} & 35.44 & 48.58 & 57.64 & 25.38 & 51.59 & 40.21 \\
& S.-S12-DINO & 81.72 & 49.37 & \textcolor{blue}{36.18} & 88.61 & 51.15 & 34.81 & 48.66 & 56.47 & 25.62 & 50.56 & 41.23 \\
& S.-S12-MAE  & \underline{81.91} & 49.90 & 32.03 & 87.79 & 51.92 & 34.08 & 45.80 & 57.13 & 24.69 & 50.44 & 41.07 \\
\midrule
\multirow{4}{*}{\textbf{Multimodal}}
& RemoteCLIP  & 76.59 & \underline{60.00} & 18.23 & 74.26 & {69.19} & 31.78 & \textcolor{blue}{52.05} & 57.76 & 25.12 & \textcolor{blue}{57.41} & {49.79} \\
& CROMA       & \textcolor{blue}{82.42} & \textcolor{blue}{67.55} & 32.32 & {90.89} & 51.83 & \textcolor{blue}{38.29} & 49.38 & 59.28 & 25.65 & \underline{53.27} & 36.81 \\
& DOFA        & 80.63 & 59.58 & 30.02 & 89.37 & 43.18 & {39.29} & \underline{51.33} & \textcolor{blue}{61.84} & \textcolor{blue}{27.07} & {59.64} & \textcolor{blue}{42.81} \\
& TerraMind   & \textcolor{blue}{82.42} & {69.52} & {40.51} & \textcolor{blue}{90.62} & \textcolor{blue}{59.72} & \underline{37.87} & {55.80} & \underline{60.61} & {28.12} & - & - \\
\bottomrule
\end{tabular}
\end{table*}

\begin{table*}
\centering
\caption{A concise comparison of existing multimodal foundation models in RS, categorized by modality type (e.g., Time-series, Geo-Vision, Homogeneous, Optical-SAR, Heterogeneous, Vision-LLM, Vision-Text, and Vision-Audio). The table details their corresponding publications, pretrained architectures, downstream applications, and code availability. CL: Contrastive Learning, MAE: Msked Autoencoders, MIM: Masked Image Modeling, Diff: Diffusion Model, JEPA: Joint Embedding Predictive Architecture, SC: Scene Classification, SS: Semantic Segmentation, RE: Regression, CD: Change Detection, OD: Object Detection, SR: Super-resolution, IG: Image Generation, VP: Visual Prediction, VG: Visual Grounding, OT: Object Tracking, IR: Image Retrieval, GM: Geometric Measurement, PS: Pan-sharpening, REC: Referring Expression Comprehension, TRE: Temporal Referring Expression, RES: Referring Expression Segmentation, VQA: Visual Question Answering, CR: Cloud Removal}
\resizebox{\textwidth}{!}{
\begin{tabular}{c|c|c|c|c|c}
\toprule[1.5pt]
Modality & Models & Publications & Architectures & Downstreams & Codes\\
\hline\hline
\addlinespace[2pt]
\multirow{5}{*}{Time-series}
& \href{https://openaccess.thecvf.com/content/ICCV2021/html/Manas_Seasonal_Contrast_Unsupervised_Pre-Training_From_Uncurated_Remote_Sensing_Data_ICCV_2021_paper.html}{SeCo}~\cite{manas2021seasonal}
& ICCV'21 & CL & SC, CD & \href{https://github.com/ServiceNow/seasonal-contrast}{https://github.com/ServiceNow/seasonal-contrast} \\
& \href{https://openaccess.thecvf.com/content/CVPR2023/html/Mall_Change-Aware_Sampling_and_Contrastive_Learning_for_Satellite_Images_CVPR_2023_paper.html}{CaCo}~\cite{mall2023change}
& CVPR'23 & CL & SC,SS, CD & \href{https://github.com/utkarshmall13/CACo}{https://github.com/utkarshmall13/CACo} \\
& \href{https://ieeexplore.ieee.org/abstract/document/10254320}{RingMo-Sense}~\cite{yao2023ringmo}
& TGRS'23 & MAE & OD,VP,OT & \href{https://github.com/}{https://github.com/}\\
& \href{https://ieeexplore.ieee.org/document/10414422}{U-BARN}~\cite{dumeur2024self}
& JSTARS'24 & MAE & SS & \href{https://src.koda.cnrs.fr/iris.dumeur/ssl_ubarn}{https://src.koda.cnrs.fr/iris.dumeur/ssl\_ubarn} \\
& \href{https://arxiv.org/abs/2412.02732}{Prithvi2.0}~\cite{szwarcman2024prithvi}
& Arxiv'24 & MAE & SC, SS, VP, GM & \href{https://github.com/NASA-IMPACT/Prithvi-EO-2.0}{https://github.com/NASA-IMPACT/Prithvi-EO-2.0} \\
\addlinespace[2pt]
\hline\hline
\addlinespace[2pt]
\multirow{5}{*}{Geo-Vision}
& \href{https://openaccess.thecvf.com/content/ICCV2021/html/Ayush_Geography-Aware_Self-Supervised_Learning_ICCV_2021_paper.html}{GASSL}~\cite{ayush2021geography}
& ICCV'21 & CL & SC, SS, OD & \href{https://github.com/sustainlab-group/geography-aware-ssl}{https://github.com/sustainlab-group/geography-aware-ssl} \\
& \href{https://ieeexplore.ieee.org/abstract/document/9559903}{GeoKR}~\cite{li2021geographical}
& TGRS'21 & CL & SC,SS, OD & \href{https://ieeexplore.ieee.org/abstract/document/9559903}{https://ieeexplore.ieee.org/abstract/document/9559903} \\
& \href{https://www.mdpi.com/2072-4292/14/22/5675\#}{CSPT}~\cite{zhang2022consecutive}
& RS'22 & MAE & SC, OD & \href{https://github.com/ZhAnGToNG1/transfer_learning_cspt}{https://github.com/ZhAnGToNG1/transfer\_learning\_cspt} \\
& \href{https://ieeexplore.ieee.org/document/10400411}{GERSP}~\cite{huang2024generic}
& TGRS'24 & CL & SS, IC, OD & \href{https://github.com/floatingstarZ/GeRSP}{https://github.com/floatingstarZ/GeRSP} \\
& \href{https://ieeexplore.ieee.org/document/10726860}{SoftCon}~\cite{wang2024multi}
& TGRS'24 &CL & SC, SS, CD & \href{https://github.com/zhu-xlab/softcon?tab=readme-ov-file}{https://github.com/zhu-xlab/softcon?tab=readme-ov-file} \\
\addlinespace[2pt]
\hline\hline
\addlinespace[2pt]
\multirow{4}{*}{Homogeneous}
& \href{https://arxiv.org/abs/2312.02199}{Usat}~\cite{irvin2023usat}
& Arxiv'23 & MAE & SC & \href{https://github.com/stanfordmlgroup/USat}{https://github.com/stanfordmlgroup/USat} \\
& \href{https://arxiv.org/abs/2408.11000}{SenPa-MAE}~\cite{prexl2024senpa}
& {GCPR'25} & MAE & SS & \href{https://github.com/JonathanPrexl/SenPa-MAE}{https://github.com/JonathanPrexl/SenPa-MAE} \\
& \href{https://arxiv.org/abs/2410.22629}{CrossEarth}~\cite{gong2024crossearth}
& Arxiv'24 & Supervised & SS & \href{https://github.com/Cuzyoung/CrossEarth}{https://github.com/Cuzyoung/CrossEarth} \\
& \href{https://arxiv.org/abs/2412.14123}{AnySat}~\cite{astruc2024anysat}
& CVPR'25 & JEPA & SC, SS & \href{https://github.com/gastruc/AnySat}{https://github.com/gastruc/AnySat} \\
\addlinespace[2pt]
\hline\hline
\addlinespace[2pt]
\multirow{7}{*}{Optical-SAR}
& \href{https://ieeexplore.ieee.org/abstract/document/9844015}{RingMo}~\cite{sun2022ringmo}
& TGRS'22 & MIM & SC, SS, CD, OD & \href{https://github.com/comeony/RingMo}{https://github.com/comeony/RingMo} \\
& \href{https://ieeexplore.ieee.org/document/9866058}{SatViT}~\cite{fuller2022satvit}
& GRSL'22 & MAE & SC & \href{https://github.com/antofuller/SatViT}{https://github.com/antofuller/SatViT} \\
& \href{https://ieeexplore.ieee.org/document/9883983}{DINO-MM}~\cite{wang2022self}
& IGARSS'22 & CL & SC & \href{https://github.com/zhu-xlab/DINO-MM}{https://github.com/zhu-xlab/DINO-MM} \\
& \href{https://ieeexplore.ieee.org/abstract/document/9880533}{RS-BYOL}~\cite{jain2022self}
& JSTARS'22 & CL & SC, SS & \href{https://github.com/}{https://github.com/} \\
& \href{https://proceedings.neurips.cc/paper_files/paper/2023/file/11822e84689e631615199db3b75cd0e4-Paper-Conference.pdf}{CROMA}~\cite{fuller2024croma}
& NeurIPS'23 & CL\&MAE & SC, SS & \href{https://github.com/antofuller/CROMA}{https://github.com/antofuller/CROMA}\\
& \href{https://arxiv.org/abs/2410.11160}{MANet}~\cite{ma2024manet}
& Arxiv'24 & Supervised & SS & \href{https://github.com/sstary/SSRS}{https://github.com/sstary/SSRS}\\
& \href{https://ieeexplore.ieee.org/document/10798628}{{CSMAE}}~\cite{hackstein2024exploring}
& {TGRS'25} & {MAE} & {IR} & \href{https://github.com/jakhac/CSMAE}{{https://github.com/jakhac/CSMAE}}\\
\addlinespace[2pt]
\hline\hline
\addlinespace[2pt]
\multirow{7}{*}{Heterogeneous}
& \href{https://openaccess.thecvf.com/content/ICCV2023/papers/Bastani_SatlasPretrain_A_Large-Scale_Dataset_for_Remote_Sensing_Image_Understanding_ICCV_2023_paper.pdf}{SatLas}~\cite{bastani2023satlaspretrain}
& ICCV'23 & Supervised & SC, SS, OD, RE, GM & \href{https://github.com/allenai/satlaspretrain_models}{https://github.com/allenai/satlaspretrain\_models} \\
& \href{https://www.ecva.net/papers/eccv_2024/papers_ECCV/papers/04236.pdf}{DeCUR}~\cite{wang2024decoupling}
& ECCV'24 &CL & SC, SS & \href{https://github.com/zhu-xlab/DeCUR}{https://github.com/zhu-xlab/DeCUR} \\
& \href{https://arxiv.org/abs/2403.15356}{DOFA}~\cite{xiong2024neural}
& Arxiv'24 & MAE & SC, SS & \href{https://github.com/zhu-xlab/DOFA}{https://github.com/zhu-xlab/DOFA} \\
& \href{https://openaccess.thecvf.com/content/CVPR2024/papers/Han_Bridging_Remote_Sensors_with_Multisensor_Geospatial_Foundation_Models_CVPR_2024_paper.pdf}{msGFM}~\cite{han2024bridging}
& CVPR'24 & MAE & SC, SS, PS, CR & \href{https://github.com/boranhan/Geospatial_Foundation_Models}{https://github.com/boranhan/Geospatial\_Foundation\_Models}\\
& \href{https://www.ecva.net/papers/eccv_2024/papers_ECCV/papers/04127.pdf}{OmniSat}~\cite{astruc2025omnisat}
& ECCV'24 & CL\&MAE & SC,SS & \href{https://github.com/gastruc/OmniSat?tab=readme-ov-file}{https://github.com/gastruc/OmniSat?tab=readme-ov-file} \\
& \href{https://www.ecva.net/papers/eccv_2024/papers_ECCV/papers/08085.pdf}{MMEarth}~\cite{nedungadi2024mmearth}
& ECCV'24 & MAE & SC, SS & \href{https://github.com/vishalned/MMEarth-train}{https://github.com/vishalned/MMEarth-train}\\
& \href{https://arxiv.org/abs/2408.09085}{MM-SAM}~\cite{xiao2024segment}
& Arxiv'24 & Supervised & SS & \href{https://xiaoaoran.github.io/projects/MM-SAM}{https://xiaoaoran.github.io/projects/MM-SAM}\\
\addlinespace[2pt]
\hline\hline
\addlinespace[2pt]
\multirow{16}{*}{Vision-LLM}
& \href{https://www.sciencedirect.com/science/article/pii/S1569843223003217}{{RSGPT}}~\cite{hu2025rsgpt}
& {ISPRS'25} & {Supervised} & {VQA, IC} & \href{https://github.com/Lavender105/RSGPT}{https://github.com/Lavender105/RSGPT}\\
& \href{https://arxiv.org/abs/2311.15826}{GeoChat}~\cite{kuckreja2024geochat}
& CVPR'24 & Supervised & VQA, IC, SC, REC, VG, OC & \href{https://github.com/mbzuai-oryx/GeoChat}{https://github.com/mbzuai-oryx/GeoChat} \\
& \href{https://www.sciencedirect.com/science/article/pii/S0924271625000206}{SkyEyeGPT}~\cite{zhan2025skyeyegpt}
& ISPRS'25 &Supervised & VQA, IC, REC, REG & \href{https://github.com/ZhanYang-nwpu/SkyEyeGPT}{https://github.com/ZhanYang-nwpu/SkyEyeGPT} \\
& \href{https://arxiv.org/pdf/2403.20213v3}{VHM}~\cite{pang2024vhm}
& AAAI'25 & Supervised & SC, VQA, IC, REC, REG, OC, GM & \href{https://github.com/opendatalab/VHM}{https://github.com/opendatalab/VHM} \\
& \href{https://www.mdpi.com/2072-4292/16/9/1477}{RS-LLaVA}~\cite{bazi2024rs}
& RS'24 & Supervised & VQA, IC & \href{https://github.com/BigData-KSU/RS-LLaVA?tab=readme-ov-file}{https://github.com/BigData-KSU/RS-LLaVA?tab=readme-ov-file} \\
& \href{https://arxiv.org/pdf/2402.02544v2}{LHRS-Bot}~\cite{muhtar2024lhrs}
& ECCV'24 & Supervised & SC, VQA, IC, REC& \href{https://github.com/NJU-LHRS/LHRS-Bot}{https://github.com/NJU-LHRS/LHRS-Bot} \\
& \href{https://ieeexplore.ieee.org/document/10547418}{EarthGPT}~\cite{zhang2024earthgpt}
& TGRS'24 &Supervised & SC, VQA, IC, REC, REG & \href{https://github.com/wivizhang/EarthGPT}{https://github.com/wivizhang/EarthGPT} \\
& \href{https://arxiv.org/abs/2406.10100}{SkysenseGPT}~\cite{luo2024skysensegpt}
& Arxiv'24 & Supervised & VQA, IC, REC, REG & \href{https://github.com/Luo-Z13/SkySenseGPT}{https://github.com/Luo-Z13/SkySenseGPT} \\
& \href{https://arxiv.org/abs/2406.07089}{RS-Agent}~\cite{xu2024rs}
& Arxiv'24 & Agent & VQA, OC, REG & \href{https://github.com/IntelliSensing/IntelliSensing.github.io}{https://github.com/IntelliSensing/IntelliSensing.github.io} \\
& \href{https://arxiv.org/abs/2407.13596}{EarthMarker}~\cite{zhang2024earthmarker}
& TGRS'24 &Supervised & VQA, SC, IC, VG, REC& \href{https://github.com/wivizhang/earthmarker}{https://github.com/wivizhang/earthmarker} \\
& \href{https://arxiv.org/abs/2411.09301}{LHRS-Bot-Nova}~\cite{li2024lhrs}
& {ISPRS'25} & Supervised & SC, VQA, IC, OD, VG, REG & \href{https://github.com/NJU-LHRS/LHRS-Bot/tree/nova}{https://github.com/NJU-LHRS/LHRS-Bot/tree/nova} \\
& \href{https://arxiv.org/html/2412.20742v1}{UniRS}~\cite{li2024unirs}
& Arxiv'24 & Supervised & VQA, IC, SC, TRE & \href{https://github.com/IntelliSensing/UniRS}{https://github.com/IntelliSensing/UniRS} \\
& \href{https://ieeexplore.ieee.org/document/10591792}{ChangeAgent}~\cite{liu2024change}
& TGRS'24 & Supervised & CD, TRE & \href{https://github.com/Chen-Yang-Liu/Change-Agent}{https://github.com/Chen-Yang-Liu/Change-Agent} \\
& \href{https://arxiv.org/abs/2412.05679}{RSUniVLM}~\cite{liu2024rsunivlm}
& Arxiv'24 & Supervised & VQA, SC, VG, CD, RES, TRE & \href{https://github.com/xuliu-cyber/RSUniVLM}{https://github.com/xuliu-cyber/RSUniVLM} \\
& \href{https://openreview.net/forum?id=pZz0nOroGv}{TEOChat}~\cite{irvin2025teochat}
& ICLR'25 &Supervised &VQA,REC,TRE & \href{https://github.com/ermongroup/TEOChat}{https://github.com/ermongroup/TEOChat} \\
& \href{https://arxiv.org/abs/2501.13925}{GeoPixel}~\cite{shabbir2025geopixel}
& {ICML'25} & Supervised & RES, VQA, VG, OC & \href{https://github.com/mbzuai-oryx/GeoPixel}{https://github.com/mbzuai-oryx/GeoPixel} \\
\addlinespace[2pt]
\hline\hline
\addlinespace[2pt]
\multirow{11}{*}{Vision-Text}
& \href{https://www.sciencedirect.com/science/article/pii/S1569843223003217}{{RS-CLIP}}~\cite{li2023rs}
& {JAG'23} & {CL} & {SC} & \href{https://github.com/lx709/RS-CLIP}{https://github.com/lx709/RS-CLIP} \\
& \href{https://ieeexplore.ieee.org/document/10504785}{RemoteCLIP}~\cite{liu2024remoteclip}
& TGRS'24 &CL & SC, IR, OC & \href{https://github.com/ChenDelong1999/RemoteCLIP}{https://github.com/ChenDelong1999/RemoteCLIP} \\
& \href{https://ieeexplore.ieee.org/document/10679571}{GeoRSCLIP}~\cite{zhang2024rs5m}
& TGRS'24 & CL & SC, IR & \href{https://github.com/om-ai-lab/RS5M}{https://github.com/om-ai-lab/RS5M} \\
& \href{https://proceedings.neurips.cc/paper_files/paper/2023/file/1b57aaddf85ab01a2445a79c9edc1f4b-Paper-Conference.pdf}{GeoCLIP}~\cite{vivanco2023geoclip}
& NeurIPS'23 & CL & SC, GM, IR & \href{https://vicentevivan.github.io/GeoCLIP/}{https://vicentevivan.github.io/GeoCLIP/} \\
& \href{https://arxiv.org/abs/2311.17179}{SatCLIP}~\cite{klemmer2023satclip}
& {AAAI'25} & CL & SC, GM, IR & \href{https://github.com/microsoft/satclip}{https://github.com/microsoft/satclip} \\
& \href{https://ojs.aaai.org/index.php/AAAI/article/view/28393}{SkyScript}~\cite{wang2024skyscript}
& AAAI'24 & CL & SC, IR & \href{https://github.com/wangzhecheng/SkyScript}{https://github.com/wangzhecheng/SkyScript} \\
& \href{https://www.ecva.net/papers/eccv_2024/papers_ECCV/papers/01738.pdf}{GeoText}~\cite{chu2024towards}
& ECCV'24 & Supervised & IR, VG, OC & \href{https://multimodalgeo.github.io/GeoText/}{https://multimodalgeo.github.io/GeoText/}\\
& \href{https://dl.acm.org/doi/abs/10.1145/3687123.3698287}{Text2Seg}~\cite{zhang2024text2seg}
& SIGSPATIAL-W'24 & CL & SS & \href{https://github.com/Douglas2Code/Text2Seg}{https://github.com/Douglas2Code/Text2Seg}\\
& \href{https://openreview.net/pdf?id=I5webNFDgQ}{DiffusionSat}~\cite{khanna2024diffusionsat}
& ICLR'24 &Diff & IG, SR & \href{https://github.com/samar-khanna/DiffusionSat}{https://github.com/samar-khanna/DiffusionSat} \\
& \href{https://ieeexplore.ieee.org/document/10663449}{CRS-Diff}~\cite{tang2403crs}
& TGRS'24 & Diff & IG & \href{https://github.com/Sonettoo/CRS-Diff}{https://github.com/Sonettoo/CRS-Diff} \\
& \href{https://arxiv.org/abs/2501.00895}{Text2Earth}~\cite{liu2025text2earth}
& {GRSM'25} & Diff & IG, IR & \href{https://chen-yang-liu.github.io/Text2Earth/}{https://chen-yang-liu.github.io/Text2Earth} \\
\addlinespace[2pt]
\hline\hline
\addlinespace[2pt]
Vision-Audio
& \href{https://www.sciencedirect.com/science/article/pii/S1569843222003181}{Sounding Earth}~\cite{heidler2023self}
& JAG'23 & CL & SC, SS, IR & \href{https://github.com/khdlr/SoundingEarth}{https://github.com/khdlr/SoundingEarth}\\
\bottomrule[1.5pt]
\end{tabular}
}
\label{tab:multimodal}
\end{table*}

\subsection{A Concise Summary of Foundation Models in RS}
We summarize unimodal and multimodal RS foundation models, detailing their corresponding publications, pretrained architectures, downstream applications, and code availability. Tables \ref{tab:unimodal} and \ref{tab:multimodal} categorize these models by unimodality and multimodality, respectively, providing insights into their architectures, practical applications, and available implementations to help readers effectively differentiate and utilize existing foundation models.

{We further demonstrate the performance gains of multimodal over unimodal foundation models in representative RS tasks, as summarized in Table~\ref{tab:multitask_performance}. More specifically, multimodal pretrained models consistently achieve superior results across classification, segmentation, and change detection tasks. To evaluate this improvement, we adopt PANGAEA-Bench, a comprehensive benchmark comprising 11 datasets covering semantic segmentation, change detection, and regression. These datasets span diverse temporal settings (mono-, bi-, and multi-temporal), global geographic distributions, spatial resolutions from 1.5 to 30 m per pixel, and heterogeneous sensor modalities (RGB, spectral, SAR). This benchmark thus provides a rigorous and unified framework for assessing RS foundation models. As shown in Table~\ref{tab:multitask_performance}, multimodal models dominate top-performing positions across nearly all tasks. In particular, on datasets such as CTM-SS, they deliver substantial gains over unimodal models, highlighting their superior generalization and effectiveness across RS applications.}

\section{How to Use Pretrained Foundation Models in RS Downstream Tasks: A Tutorial}
\begin{figure*}[!t]
   \centering
		\includegraphics[width=1.0\textwidth]{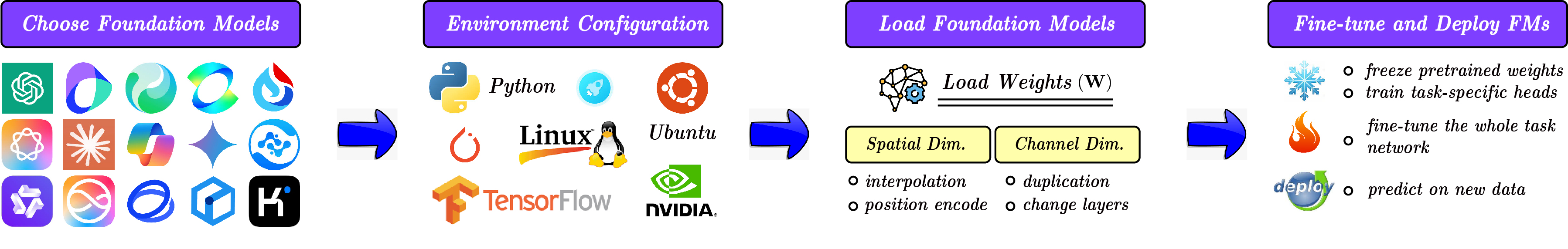}
    \caption{A step-by-step workflow for applying pretrained foundation models in RS downstream tasks. This tutorial-like process begins with model selection, followed by environment configuration and model loading, and concludes with fine-tuning and deploying the chosen foundation models for specific RS applications.}
\label{fig:tutorial}
\end{figure*}

Pretrained foundation models in RS offer robust capabilities for downstream tasks by leveraging extensive pretraining on massive datasets. Efficiently applying these models to various tasks is a critical focus in RS research and practice. This chapter provides a step-by-step guide covering model selection, environment setup, model loading, fine-tuning, and deployment, offering practical methods for utilizing RS foundation models. A detailed workflow for these steps is shown in Fig. \ref{fig:tutorial}.

\subsection{Choose Right RS Foundation Models}
 Different RS foundation models exhibit unique characteristics in terms of pretraining data, network architecture, and model features. Therefore, selecting the right model requires careful evaluation based on the needs of the downstream task.

First, consider the source and characteristics of the pretraining data, such as the type of sensor used, the satellite platform, image dimensions, channel count, and data format (e.g., integer or floating-point). These factors help assess the alignment between the model's pretraining data and the data requirements of the downstream task.

Second, model architecture plays a crucial role. Evaluate whether the model employs a Transformer, a convolutional network, or a hybrid backbone, along with the parameter count and computational complexity, to ensure compatibility with the task's resource constraints. Additionally, assess whether the architecture meets specific requirements for lightweight deployment or real-time performance.

Finally, one should consider the model's unique features, such as its ability to integrate multimodal data or incorporate temporal information. These features can significantly enhance performance in specialized tasks.

\subsection{Model Configuration}
After selecting a foundation model, the next step is configuring the corresponding runtime environment based on its documentation. The core goal is to ensure the libraries and frameworks the model requires are correctly set up.

Most RS foundation models rely on Python and deep learning frameworks like PyTorch. Additionally, tools for processing RS data, such as rasterio\footnote{https://github. com/mapbox/rasterio} and GDAL, are often necessary. When setting up the environment, one should pay close attention to installing framework versions compatible with the selected model. Ensure that pre-trained weights are correctly downloaded and stored in the specified locations.

Properly configuring the environment is essential to support subsequent tasks, such as model loading and fine-tuning, and ensures reliable operation throughout the workflow.

\subsection{Load Pretrained Foundation Models}
The process of loading pretrained models involves more than just importing weights; it also requires addressing challenges stemming from the unique characteristics of RS data. Variations in image dimensions and channel counts can prevent pretrained models from functioning correctly if left unaddressed.

\subsubsection{Handling Mismatches w.r.t Image Dimension}
\begin{itemize}
    \item \textbf{Issue:} Pretrained models are typically designed for fixed image sizes (e.g., $224\times 224 \times 3$), whereas the image dimensions in downstream tasks may differ (e.g., $128\times 128\times 3$).
    \item \textbf{Solutions:} (i) Image Resizing or Interpolation: Adjust the dimensions of the input images to match the requirements of the pretrained model. (ii) Positional Encoding Interpolation: For models using positional encodings (e.g., transformers), interpolate the encoding values to adapt to the new input dimensions while maintaining spatial consistency.
\end{itemize}

\subsubsection{Handling Mismatches w.r.t Channel Dimension}
\begin{itemize}
    \item \textbf{Issue:} RS data often exhibit significant variability in channel counts, such as single-channel SAR images or spectral images with dozens or hundreds of channels.
    \item \textbf{Solutions:} (i) Channel Duplication or principal component analysis (PCA) Dimensionality Reduction: Replicate channels or reduce the input dimensionality using PCA to align with the model's requirements. (ii) Adding Convolutional Layers: Introduce a convolutional layer at the input stage to transform the channel count to match the pretrained model. (iii) Modifying the Input Layer: Directly adjust the model's input layer to accommodate the new channel count, ensuring compatibility with the specific data characteristics.
\end{itemize}

\subsection{Fine-tune and Deploy Foundation Models}
After loading the model, fine-tuning is necessary to adapt it to specific tasks. Fine-tuning strategies can range from freezing certain parameters and training only particular modules, such as the input or output layers, to fully fine-tuning the entire model.

\subsubsection{Choosing a Fine-tuning Strategy}
\begin{itemize}
    \item \textbf{Resource-aware Fine-tuning:} For scenarios with limited computational resources, freeze the pretrained parameters and adjust only task-specific modules to reduce computational overhead.
    \item \textbf{Comprehensive Fine-tuning:} For tasks requiring high precision or when resources are abundant, fine-tune the entire model to optimize it fully for the downstream task.
\end{itemize}

\subsubsection{Deploying the Fine-tuned Model}
Once fine-tuning is complete, the model can be deployed to make predictions on new data. The fine-tuned model leverages the generalization capabilities from the pretraining phase and the task-specific optimizations from fine-tuning, often achieving superior accuracy and robustness in practical applications.

\subsection{Tutorial Extension: Vision-Language Foundation Models for RS}
\label{sec:vl_tutorial}
\textbf{Why this extension?} The above tutorial focuses on vision-only foundation models. However, many real-world RS applications require joint reasoning over imagery and textual/metadata information (e.g., scene descriptions, sensor metadata, temporal or geospatial attributes). To address the reviewer's concern, we provide a practical, step-by-step guide for adopting \emph{vision-language foundation models} (VLFMs) in RS workflows. This section mirrors the previous pipeline while emphasizing multimodal specifics.

\subsubsection{Step 1: Clarify Task Type and Data Pairing}
\begin{itemize}
    \item \textbf{Task taxonomy:} Identify whether the downstream task is \emph{image-to-text} (captioning, report generation), \emph{text-to-image retrieval}, \emph{visual question answering (VQA)}, \emph{instruction following / dialogue}, or \emph{multimodal classification/detection} with textual prompts.
    \item \textbf{Paired data construction:} RS corpora rarely come with rich human-written captions. Consider (i) leveraging existing metadata (sensor name, acquisition time, geolocation, land-cover labels) to auto-generate weak captions; (ii) using large language models (LLMs) to refine noisy metadata into natural language; and (iii) adopting teacher models (e.g., generic VLFMs) to pseudo-caption images, followed by human filtering for critical datasets.
    \item \textbf{Granularity of text:} Decide if you need per-image, per-tile, or per-object text. Fine-grained supervision often improves VQA/grounding, but increases annotation cost.
\end{itemize}

\subsubsection{Step 2: Choose/Configure an RS Vision--Language FM}
\begin{itemize}
    \item \textbf{Model families:}
    \begin{enumerate}
        \item \emph{Encoder-decoder style} (e.g., BLIP-2-like): a frozen vision encoder, a lightweight connector (Q-Former / MLP / cross-attn), and a frozen or partially tunable LLM.
        \item \emph{Unified transformers} (single backbone for both modalities) that jointly encode tokens from vision and text.
        \item \emph{RS-specific VLFMs} (e.g., GeoChat-like), already adapted to geographic semantics and RS vocabulary.
    \end{enumerate}
    \item \textbf{Connector design:} Ensure the connector can handle RS-specific vision tokens (e.g., many bands, large tiles). For spectrum-rich data, prepend band \emph{text tokens} (``Band-11 (SWIR 1.61 µm)'') or wavelength descriptors to help the language side understand spectral semantics.
    \item \textbf{Tokenizer and vocabulary:} Confirm that sensor names, land-cover classes, and geospatial terms exist in the tokenizer; if not, extend the vocabulary or map to existing tokens via aliases.
\end{itemize}

\subsubsection{Step 3: Load and Align Multimodal Weights}
\begin{itemize}
    \item \textbf{Vision side mismatches:} Reuse the image-dimension/channel strategies from Section~IV (resizing, positional encoding interpolation, input adapters). For spectral inputs, consider (i) spectral grouping or 1$\times$1 conv, \emph{and simultaneously} (ii) pass textual descriptors of dropped/merged bands to the language side to compensate for lost information.
    \item \textbf{Text side initialization:} Load the LLM tokenizer and weights first; then initialize or map the connector parameters. If the connector expects a fixed number of visual tokens, interpolate or sparsely sample patch tokens to match.
    \item \textbf{Cross-modal alignment heads:} Some checkpoints with image--text contrastive or matching heads. Verify dimension consistency when swapping encoders or tokenizers.
\end{itemize}

\subsubsection{Step 4: Multimodal Fine-tuning Strategies}
\begin{itemize}
    \item \textbf{Lightweight adaptation (resource-aware):} Freeze both the vision encoder and LLM, and fine-tune only (i) the connector (e.g., Q-Former), (ii) LoRA/adapter layers in the LLM, and/or (iii) a shallow projection for spectral channels. This is often enough for retrieval and captioning.
    \item \textbf{Instruction tuning:} For VQA or dialogue-style tasks, curate instruction--response pairs (questions about RS scenes, answers grounded in imagery). Apply supervised fine-tuning (SFT), then optionally preference optimization (DPO/RLHF) if human preference data is available.
    \item \textbf{Multi-task schedules:} Combine image--text contrastive loss, image-conditioned language modeling, and multilingual targets (e.g., English/Chinese) to bolster generalization.
\end{itemize}

\subsubsection{Step 5: Deployment and Prompting}
\begin{itemize}
    \item \textbf{Prompt design:} Provide structured prompts that expose RS context: ``You are an RS analyst. Given a Sentinel-2 image (10 m GSD, bands: B2, B3, B4, B8), answer the question.''. Such explicit cues reduce hallucination.
    \item \textbf{Inference efficiency:} For large LLMs, distill to smaller language heads or cache key-value states for repeated queries. On the vision side, precompute image embeddings and index them for retrieval tasks.
    \item \textbf{Evaluation:} Beyond standard NLP metrics (BLEU, CIDEr), adopt RS-aware measures (land-cover accuracy, geospatial grounding accuracy) and human expert evaluation for safety-critical tasks.
\end{itemize}

\noindent  \textbf{Takeaway.} Integrating VLFMs into RS pipelines requires (1) constructing or curating reliable image-text pairs, (2) choosing an appropriate multimodal architecture/connector, (3) carefully aligning spectral/spatial tokens with textual descriptors, and (4) adopting resource-sensitive fine-tuning and prompt strategies. Following these steps enables robust cross-modal understanding and interaction in practical RS applications.

\subsection{Computational Resources for Foundation Models}
{To help readers or potential researchers in selecting appropriate foundation models based on their computational resources, we summarize representative backbone architectures for remote sensing foundation models in Table \ref{tab:comp_resources}. Unless otherwise noted, all values are computed (or re-computed) under the following consistent assumptions: \emph{(i)} a single forward pass, \emph{(ii)} FP32 precision, \emph{(iii)} batch size of 1, and \emph{(iv)} an input resolution of $3\times224\times224$ for CNN-, ViT-, and Swin-based models. For SAM variants, FLOPs are reported using their original input setting ($1024\times1024$) as specified in the SAM paper. FLOPs denote floating-point operations per forward pass; when multiply-accumulate operations (MACs) are reported, we convert them to FLOPs by multiplying by two. Parameter counts (``Params'') refer to the total number of learnable weights, while ``Inference Memory'' indicates an empirical range measured on NVIDIA GPUs, subject to variation across drivers, PyTorch versions, and activation checkpointing. For widely studied models (e.g., ResNet-50, ViT-B/L/H, Swin-T/B/L), we adopt official or commonly accepted reference implementations.}

We also report the computational profiles of mainstream open-source foundation models in RS to complement the backbone-level summary, as listed in Table~\ref{tab:comp_resources}. Moreover, Table~\ref{tab:fm_complexity} presents each model’s complexity (in GMacs), number of trainable parameters, and total parameters, as provided by the \textsc{PANGAEA}-Bench release. Unless otherwise noted, all GMacs are measured as multiply-accumulate operations per single forward pass (1~MAC = 2~FLOPs) under the benchmark’s default input configuration. ``Trainable Params'' denote the subset of weights updated during fine-tuning for downstream tasks. This table is intended to help readers balance potential accuracy gains against computational costs when selecting among current mainstream foundation models in RS.

\begin{table*}[!t]
\centering
\renewcommand{\arraystretch}{1.1}
\caption{Computational resources of representative backbone architectures for foundation models in RS.}
\label{tab:comp_resources}
\begin{tabular}{llccc}
\toprule
\textbf{Model} & \textbf{Architecture} & \textbf{Params (M)} & \textbf{FLOPs (G)} & \textbf{Inference Memory (GB)} \\
\midrule
ResNet-50  & ResNet-50            & 25.6  & 4.1   & 1--2  \\
ViT-Base   & ViT-B/16             & 86.0  & 17.6  & 2--4  \\
ViT-Large  & ViT-L/16             & 304.0 & 61.6  & 4--8  \\
ViT-Huge   & ViT-H/14             & 632.0 & 190.7 & 8--12 \\
SAM-B      & ViT-B/16 (SAM)       & 91.0  & 22.0  & 2--4  \\
SAM-L      & ViT-L/16 (SAM)       & 308.0 & 62.0  & 4--8  \\
SAM-H      & ViT-H/14 (SAM)       & 641.0 & 200.0 & 8--12 \\
Swin-T     & Swin-T               & 28.3  & 4.5   & 1--2  \\
Swin-B     & Swin-B               & 87.8  & 15.4  & 2--4  \\
Swin-L     & Swin-L               & 197.3 & 34.5  & 4--6  \\
\bottomrule
\end{tabular}
\end{table*}

\begin{table*}[!t]
\centering
\renewcommand{\arraystretch}{1.1}
\caption{Computational statistics of representative foundation models in RS.}
\label{tab:fm_complexity}
\begin{tabular}{lccc}
\toprule
\textbf{Model} & {\textbf{FLOPs (G)}} & \textbf{Trainable Params (M)} & \textbf{Total Params (M)} \\
\midrule
CROMA        & {278.46} & 46.95 & 350.0 \\
DOFA         & {130.04} & 39.35 & 150.9 \\
GFM-Swin     &  {38.10} & 33.71 & 120.4 \\
Prithvi      & {129.70} & 39.34 & 125.7 \\
RemoteCLIP   &  {35.14} & 39.35 &  16.8 \\
SatlasNet    &  {33.40} & 33.25 & 121.2 \\
Scale-MAE    & {207.54} & 46.95 & 350.1 \\
SpectralGPT  & {761.38} & 164.40& 249.8 \\
S12-MoCo     &  {93.80} & 30.89 &  53.5 \\
S12-DINO     &  {95.86} & 30.89 &  53.5 \\
S12-MAE      &  {93.80} & 30.89 &  53.5 \\
S12-Data2Vec &  {93.78} & 30.89 &  53.5 \\
\bottomrule
\end{tabular}
\end{table*}

\section{Conclusion}
{Foundation models have revolutionized the field of remote sensing (RS) by enabling scalable and transferable learning from large, heterogeneous data sources. Early RS foundation models predominantly focused on unimodal representations, learning generic features within individual sensing modalities. More recently, this paradigm has evolved toward multimodal foundation models that jointly exploit complementary information across optical RGB, SAR, spectral, LiDAR, and auxiliary data sources. Leveraging large-scale pretraining and task-adaptive fine-tuning, these models establish a unified and robust framework for a wide range of downstream tasks, including image classification, object detection, semantic segmentation, change detection, and beyond.}

{In addition, this review incorporates a tutorial-oriented perspective, systematically outlining the essential steps for effectively utilizing pretrained foundation models in RS applications, including model selection, environment configuration, model loading, fine-tuning, and deployment. Each step requires careful consideration of the unique characteristics of RS data, computational constraints, and domain-specific challenges to ensure robust performance and practical adaptability.}

\section{Future Directions and Challenges}
Despite significant progress, the development of foundation models in RS presents several open challenges that warrant further investigation:
\begin{itemize}
    \item \textbf{Scaling Laws and Model Growth:} Understanding whether scaling laws hold for foundation models in RS is crucial for guiding model development, optimizing data requirements, and balancing computational efficiency with performance.
    \item \textbf{Defining Generality and Model Criteria:} Future research should establish clear criteria for what constitutes a general or sufficiently large RS foundation model and clarify whether such models must be self-supervised.
    \item \textbf{Comparative Analysis of Pretrained Models:} A deeper investigation is needed into why, in many cases, random weights or models pretrained on other domains yield results comparable to existing RS-specific foundation models.
    \item \textbf{Catastrophic Forgetting and Knowledge Retention:} Addressing the issue of catastrophic forgetting in trained RS foundation models is essential for ensuring long-term adaptability and efficient continuous learning.
    \item \textbf{Data Management and Scalability:} Managing large-scale RS data, including preprocessing, annotation, and efficient storage, remains a bottleneck. Future research should explore standardized frameworks and strategies to facilitate scalable development of RS foundation models.
    \item \textbf{Multimodal Adaptability and Open Collaboration:} Enhancing model adaptability to diverse RS data modalities and fostering collaborative efforts in sharing pretrained models, datasets, and benchmarks will accelerate innovation across the community.
    {\item \textbf{Robustness and Uncertainty:} Future RS foundation models should move beyond post-hoc robustness evaluation (e.g., REOBench \cite{li2025reobench}) toward robustness-aware and uncertainty-aware pretraining, explicitly modeling sensor noise, domain shifts, and extreme conditions. Integrating uncertainty quantification (e.g., probabilistic representations or Bayesian-inspired learning) will be crucial to enable reliable decision-making and risk-aware deployment in real-world EO applications.}
\end{itemize}

By tackling these challenges, foundation models will continue to drive advancements in RS, enabling more accurate environmental monitoring, sustainable resource management, and deeper insights into the Earth's dynamic processes.

\section{Main Abbreviation}
For clarity, we have compiled all abbreviations used throughout the paper as follows. We hope this list helps readers locate and understand them more easily.

\vspace{5pt}
\begin{supertabular}{ll}
CNNs: & convolutional neural networks\\
CROMA: & contrastive radar-optical masked autoencoders\\
CSPT: & consecutive pretraining\\
CWSAM: & class-wise SAM\\
DOFA: & dynamic one for all\\
EO: & Earth observation\\
FG-MAE: & feature guided masked autoencoders\\
GASSL: & geography-aware self-supervised learning\\
GDAL: & geospatial data abstraction library\\
GeoKR: & geographical knowledge-driven representation\\
& learning\\
GFM: & geospatial foundation models\\
GRAFT: & ground remote alignment for training\\
GSD: & ground sampling distance\\
HOGs: & histograms of oriented gradients\\
LoRA: & low-rank adaptation\\
LMM: & large language models\\
MAE: & masked autoencoders\\
MAF-SAM: & multistage adaptation fine-tuning segment\\
& anything model\\
MANet: & multimodal adapter-based network\\
MATTER: & material and texture representation learning\\
MLP: & multilayer perception\\
msGFM: & multisensor geospatial foundation model\\
omniSat: & omni satellite\\
OFA-Net: & one for all network\\
RingMo: & remote sensing foundation model\\
RNNs: & recurrent neural networks\\
RS: & remote sensing\\
RS-BYOL: & remote sensing bootstrap your own latent\\
RS-DFM: & remote sensing distributed foundation model\\
RVSA: & rotated varied-size attention\\
S2MAE: & spatial-spectral masked autoencoders\\
SatMAE: & satellite masked autoencoders\\
SatViT: & satellite vision transformer\\
SpectralGPT: & spectral generative pretrained transformer\\
SAM: & segment anything model\\
SAR: & synthetic aperture radar\\
SARATR-X: & SAR automatic target recognition\\
SAR-JEPA: & SAR joint-embedding predictive architecture\\
SeaMo: & seasonal and multimodal foundation model\\
SenPa-MAE: & sensor parameter masked autoencoders\\
SoftCon: & soft contrastive learning\\
ViTs: & vision transformers\\
\end{supertabular}

\section{Acknowledgements}
This work was supported by the National Natural Science Foundation of China under Grant 42271350 and by the International Partnership Program of the Chinese Academy of Sciences under Grant No.313GJHZ2023066FN. GCV gratefully acknowledges the support from the European Research Council (ERC) through the USMILE project (grant agreement 855187) and the European Commission through the HORIZON projects ELIAS (grant agreement 101120237) and ELLIOT (grant agreement101214398).

\bibliographystyle{ieeetr}
\bibliography{HDF_ref}

@article{li2024interpretable,
  title={Interpretable foundation models as decryptors peering into the Earth system},
  author={Li, Chenyu and Hong, Danfeng and Zhang, Bing and Liao, Tianjun and Yokoya, Naoto and Ghamisi, Pedram and Chen, Min and Wang, Lizhe and Benediktsson, Jon Atli and Chanussot, Jocelyn},
  journal={The Innovation},
  volume={5},
  number={5},
  year={2024},
  publisher={Elsevier}
}

@article{hong2024multimodal,
  title={Multimodal artificial intelligence foundation models: Unleashing the power of remote sensing big data in earth observation},
  author={Hong, Danfeng and Li, Chenyu and Zhang, Bing and Yokoya, Naoto and Benediktsson, Jon Atli and Chanussot, Jocelyn},
  journal={The Innovation Geoscience},
  volume={2},
  number={1},
  pages={100055},
  year={2024},
  publisher={The Innovation Geoscience}
}

@article{wang2024mtp,
  title={Mtp: Advancing remote sensing foundation model via multi-task pretraining},
  author={Wang, Di and Zhang, Jing and Xu, Minqiang and Liu, Lin and Wang, Dongsheng and Gao, Erzhong and Han, Chengxi and Guo, Haonan and Du, Bo and Tao, Dacheng and others},
  journal={IEEE Journal of Selected Topics in Applied Earth Observations and Remote Sensing},
  year={2024},
  publisher={IEEE}
}

@article{li2025urbansam,
  title={UrbanSAM: Learning Invariance-Inspired Adapters for Segment Anything Models in Urban Construction},
  author={Li, Chenyu and Hong, Danfeng and Zhang, Bing and Li, Yuxuan and Camps-Valls, Gustau and Zhu, Xiao Xiang and Chanussot, Jocelyn},
  journal={arXiv preprint arXiv:2502.15199},
  year={2025}
}

@inproceedings{akiva2022self,
  title={Self-supervised material and texture representation learning for remote sensing tasks},
  author={Akiva, Peri and Purri, Matthew and Leotta, Matthew},
  booktitle={Proc. CVPR},
  pages={8203--8215},
  year={2022}
}

@article{wang2022advancing,
  title={Advancing plain vision transformer toward remote sensing foundation model},
  author={Wang, Di and Zhang, Qiming and Xu, Yufei and Zhang, Jing and Du, Bo and Tao, Dacheng and Zhang, Liangpei},
  journal={IEEE Transactions on Geoscience and Remote Sensing},
  volume={61},
  pages={1--15},
  year={2022},
  publisher={IEEE}
}

@article{long2021creating,
  title={On creating benchmark dataset for aerial image interpretation: Reviews, guidances, and million-aid},
  author={Long, Yang and Xia, Gui-Song and Li, Shengyang and Yang, Wen and Yang, Michael Ying and Zhu, Xiao Xiang and Zhang, Liangpei and Li, Deren},
  journal={IEEE Journal of Selected Topics in Applied Earth Observations and Remote Sensing},
  volume={14},
  pages={4205--4230},
  year={2021},
  publisher={IEEE}
}

@inproceedings{mendieta2023towards,
  title={Towards geospatial foundation models via continual pretraining},
  author={Mendieta, Mat{\'\i}as and Han, Boran and Shi, Xingjian and Zhu, Yi and Chen, Chen},
  booktitle={Pro. ICCV},
  pages={16806--16816},
  year={2023}
}

@article{cong2022satmae,
  title={Satmae: Pre-training transformers for temporal and multi-spectral satellite imagery},
  author={Cong, Yezhen and Khanna, Samar and Meng, Chenlin and Liu, Patrick and Rozi, Erik and He, Yutong and Burke, Marshall and Lobell, David and Ermon, Stefano},
  journal={Proc. NeurIPS},
  volume={35},
  pages={197--211},
  year={2022}
}

@inproceedings{wanyan2024extending,
  title={Extending global-local view alignment for self-supervised learning with remote sensing imagery},
  author={Wanyan, Xinye and Seneviratne, Sachith and Shen, Shuchang and Kirley, Michael},
  booktitle={Proceedings of the IEEE/CVF Conference on Computer Vision and Pattern Recognition},
  pages={2443--2453},
  year={2024}
}

@article{song2024multispectral,
  title={A multispectral remote sensing crop segmentation method based on Segment Anything Model using Multi-stage Adaptation Fine-tuning},
  author={Song, Binbin and Yang, Hui and Wu, Yanlan and Zhang, Peng and Wang, Biao and Han, Guichao},
  journal={IEEE Transactions on Geoscience and Remote Sensing},
  year={2024},
  publisher={IEEE}
}

@article{ibanez2022masked,
  title={Masked auto-encoding spectral--spatial transformer for hyperspectral image classification},
  author={Ibanez, Damian and Fernandez-Beltran, Ruben and Pla, Filiberto and Yokoya, Naoto},
  journal={IEEE Transactions on Geoscience and Remote Sensing},
  volume={60},
  pages={1--14},
  year={2022},
  publisher={IEEE}
}

@inproceedings{scheibenreif2023masked,
  title={Masked vision transformers for hyperspectral image classification},
  author={Scheibenreif, Linus and Mommert, Michael and Borth, Damian},
  booktitle={Proceedings of the IEEE/CVF Conference on Computer Vision and Pattern Recognition Workshops (CVPRW)},
  pages={2166--2176},
  year={2023}
}

@article{hsu2024geospatial,
  title={Geospatial foundation models for image analysis: evaluating and enhancing NASA-IBM Prithvi’s domain adaptability},
  author={Hsu, Chia-Yu and Li, Wenwen and Wang, Sizhe},
  journal={International Journal of Geographical Information Science},
  pages={1--30},
  year={2024},
  publisher={Taylor \& Francis}
}

@article{hong2024spectralgpt,
  title={SpectralGPT: Spectral remote sensing foundation model},
  author={Hong, Danfeng and Zhang, Bing and Li, Xuyang and Li, Yuxuan and Li, Chenyu and Yao, Jing and Yokoya, Naoto and Li, Hao and Ghamisi, Pedram and Jia, Xiuping and others},
  journal={IEEE Transactions on Pattern Analysis and Machine Intelligence},
  volume={46},
  number={8},
  pages={5227--5244},
  year={2024},
  publisher={IEEE}
}

@article{li2026seamo,
  title={SeaMo: A season-aware multimodal foundation model for remote sensing},
  author={Li, Xuyang and Li, Chenyu and Vivone, Gemine and Hong, Danfeng},
  journal={Information Fusion},
  volume={125},
  pages={103334},
  year={2026},
  publisher={Elsevier}
}

@article{wang2024feature,
  title={Feature guided masked autoencoder for self-supervised learning in remote sensing},
  author={Wang, Yi and Hernandez, Hugo Hernandez and Albrecht, Conrad M and Zhu, Xiao Xiang},
  journal={IEEE Journal of Selected Topics in Applied Earth Observations and Remote Sensing},
  year={2024},
  publisher={IEEE}
}

@article{li2025saratr,
  title={SARATR-X: Towards Building A Foundation Model for SAR Target Recognition},
  author={Li, Weijie and Yang, Wei and Hou, Yuenan and Liu, Li and Liu, Yongxiang and Li, Xiang},
  journal={IEEE Transactions on Image Processing},
  year={2025},
  publisher={IEEE}
}

@article{li2024predicting,
  title={Predicting gradient is better: Exploring self-supervised learning for SAR ATR with a joint-embedding predictive architecture},
  author={Li, Weijie and Yang, Wei and Liu, Tianpeng and Hou, Yuenan and Li, Yuxuan and Liu, Zhen and Liu, Yongxiang and Liu, Li},
  journal={ISPRS Journal of Photogrammetry and Remote Sensing},
  volume={218},
  pages={326--338},
  year={2024},
  publisher={Elsevier}
}

@article{pu2024classwise,
  title={Classwise-sam-adapter: Parameter efficient fine-tuning adapts segment anything to sar domain for semantic segmentation},
  author={Pu, Xinyang and Jia, Hecheng and Zheng, Linghao and Wang, Feng and Xu, Feng},
  journal={arXiv preprint arXiv:2401.02326},
  year={2024}
}

@article{wang2024hsimae,
  title={HSIMAE: A Unified Masked Autoencoder with Large-scale Pre-training for Hyperspectral Image Classification},
  author={Wang, Yue and Wen, Ming and Zhang, Hailiang and Sun, Jinyu and Yang, Qiong and Zhang, Zhimin and Lu, Hongmei},
  journal={IEEE Journal of Selected Topics in Applied Earth Observations and Remote Sensing},
  year={2024},
  publisher={IEEE}
}

@article{braham2024spectralearth,
  title={SpectralEarth: Training Hyperspectral Foundation Models at Scale},
  author={Braham, Nassim Ait Ali and Albrecht, Conrad M and Mairal, Julien and Chanussot, Jocelyn and Wang, Yi and Zhu, Xiao Xiang},
  journal={arXiv preprint arXiv:2408.08447},
  year={2024}
}

@article{wang2024hypersigma,
  title={HyperSIGMA: Hyperspectral Intelligence Comprehension Foundation Model},
  author={Wang, Di and Hu, Meiqi and Jin, Yao and Miao, Yuchun and Yang, Jiaqi and Xu, Yichu and Qin, Xiaolei and Ma, Jiaqi and Sun, Lingyu and Li, Chenxing and others},
  journal={arXiv preprint arXiv:2406.11519},
  year={2024}
}

@article{irvin2023usat,
  title={USat: A unified self-supervised encoder for multi-sensor satellite imagery},
  author={Irvin, Jeremy and Tao, Lucas and Zhou, Joanne and Ma, Yuntao and Nashold, Langston and Liu, Benjamin and Ng, Andrew Y},
  journal={arXiv preprint arXiv:2312.02199},
  year={2023}
}

@article{wang2024rs,
  title={RS-DFM: A Remote Sensing Distributed Foundation Model for Diverse Downstream Tasks},
  author={Wang, Zhechao and Cheng, Peirui and Tian, Pengju and Wang, Yuchao and Chen, Mingxin and Duan, Shujing and Wang, Zhirui and Li, Xinming and Sun, Xian},
  journal={arXiv preprint arXiv:2406.07032},
  year={2024}
}

@article{prexl2024senpa,
  title={SenPa-MAE: Sensor Parameter Aware Masked Autoencoder for Multi-Satellite Self-Supervised Pretraining},
  author={Prexl, Jonathan and Schmitt, Michael},
  journal={arXiv preprint arXiv:2408.11000},
  year={2024}
}

@article{sun2022ringmo,
  title={RingMo: A remote sensing foundation model with masked image modeling},
  author={Sun, Xian and Wang, Peijin and Lu, Wanxuan and Zhu, Zicong and Lu, Xiaonan and He, Qibin and Li, Junxi and Rong, Xuee and Yang, Zhujun and Chang, Hao and others},
  journal={IEEE Transactions on Geoscience and Remote Sensing},
  volume={61},
  pages={1--22},
  year={2022},
  publisher={IEEE}
}

@article{fuller2022satvit,
  title={Satvit: Pretraining transformers for earth observation},
  author={Fuller, Anthony and Millard, Koreen and Green, James R},
  journal={IEEE Geoscience and Remote Sensing Letters},
  volume={19},
  pages={1--5},
  year={2022},
  publisher={IEEE}
}

@article{jain2022self,
  title={Self-supervised learning for invariant representations from multi-spectral and SAR images},
  author={Jain, Pallavi and Schoen-Phelan, Bianca and Ross, Robert},
  journal={IEEE Journal of Selected Topics in Applied Earth Observations and Remote Sensing},
  volume={15},
  pages={7797--7808},
  year={2022},
  publisher={IEEE}
}

@article{fuller2024croma,
  title={CROMA: Remote sensing representations with contrastive radar-optical masked autoencoders},
  author={Fuller, Anthony and Millard, Koreen and Green, James},
  journal={Proc. NeurIPS},
  volume={36},
  year={2024}
}

@article{ma2024manet,
  title={MANet: Fine-Tuning Segment Anything Model for Multimodal Remote Sensing Semantic Segmentation},
  author={Ma, Xianping and Zhang, Xiaokang and Pun, Man-On and Huang, Bo},
  journal={arXiv preprint arXiv:2410.11160},
  year={2024}
}

@inproceedings{han2024bridging,
  title={Bridging remote sensors with multisensor geospatial foundation models},
  author={Han, Boran and Zhang, Shuai and Shi, Xingjian and Reichstein, Markus},
  booktitle={Proc. CVPR},
  pages={27852--27862},
  year={2024}
}

@article{xiong2024one,
  title={One for all: Toward unified foundation models for Earth vision},
  author={Xiong, Zhitong and Wang, Yi and Zhang, Fahong and Zhu, Xiao Xiang},
  journal={arXiv preprint arXiv:2401.07527},
  year={2024}
}

@article{xiong2024neural,
  title={Neural plasticity-inspired foundation model for observing the Earth crossing modalities},
  author={Xiong, Zhitong and Wang, Yi and Zhang, Fahong and Stewart, Adam J and Hanna, Jo{\"e}lle and Borth, Damian and Papoutsis, Ioannis and Le Saux, Bertrand and Camps-Valls, Gustau and Zhu, Xiao Xiang},
  journal={arXiv e-prints},
  pages={arXiv--2403},
  year={2024}
}

@inproceedings{astruc2025omnisat,
  title={Omnisat: Self-supervised modality fusion for earth observation},
  author={Astruc, Guillaume and Gonthier, Nicolas and Mallet, Clement and Landrieu, Loic},
  booktitle={European Conference on Computer Vision},
  pages={409--427},
  year={2025},
  organization={Springer}
}

@article{liu2024remoteclip,
  title={Remoteclip: A vision language foundation model for remote sensing},
  author={Liu, Fan and Chen, Delong and Guan, Zhangqingyun and Zhou, Xiaocong and Zhu, Jiale and Ye, Qiaolin and Fu, Liyong and Zhou, Jun},
  journal={IEEE Transactions on Geoscience and Remote Sensing},
  year={2024},
  publisher={IEEE}
}

@article{zhang2024rs5m,
  title={RS5M and GeoRSCLIP: A large scale vision-language dataset and a large vision-language model for remote sensing},
  author={Zhang, Zilun and Zhao, Tiancheng and Guo, Yulong and Yin, Jianwei},
  journal={IEEE Transactions on Geoscience and Remote Sensing},
  year={2024},
  publisher={IEEE}
}

@article{mall2023remote,
  title={Remote sensing vision-language foundation models without annotations via ground remote alignment},
  author={Mall, Utkarsh and Phoo, Cheng Perng and Liu, Meilin Kelsey and Vondrick, Carl and Hariharan, Bharath and Bala, Kavita},
  journal={arXiv preprint arXiv:2312.06960},
  year={2023}
}

@inproceedings{wang2024skyscript,
  title={Skyscript: A large and semantically diverse vision-language dataset for remote sensing},
  author={Wang, Zhecheng and Prabha, Rajanie and Huang, Tianyuan and Wu, Jiajun and Rajagopal, Ram},
  booktitle={Proc. AAAI},
  volume={38},
  number={6},
  pages={5805--5813},
  year={2024}
}

@inproceedings{zhang2024text2seg,
  title={Text2Seg: Zero-shot Remote Sensing Image Semantic Segmentation via Text-Guided Visual Foundation Models},
  author={Zhang, Jielu and Zhou, Zhongliang and Mai, Gengchen and Hu, Mengxuan and Guan, Zihan and Li, Sheng and Mu, Lan},
  booktitle={Proc. ACM SIGSPATIAL International Workshop on AI for Geographic Knowledge Discovery},
  pages={63--66},
  year={2024}
}

@article{heidler2023self,
  title={Self-supervised audiovisual representation learning for remote sensing data},
  author={Heidler, Konrad and Mou, Lichao and Hu, Di and Jin, Pu and Li, Guangyao and Gan, Chuang and Wen, Ji-Rong and Zhu, Xiao Xiang},
  journal={International Journal of Applied Earth Observation and Geoinformation},
  volume={116},
  pages={103130},
  year={2023},
  publisher={Elsevier}
}

@inproceedings{ayush2021geography,
  title={Geography-aware self-supervised learning},
  author={Ayush, Kumar and Uzkent, Burak and Meng, Chenlin and Tanmay, Kumar and Burke, Marshall and Lobell, David and Ermon, Stefano},
  booktitle={Proc. ICCV},
  pages={10181--10190},
  year={2021}
}

@article{li2021geographical,
  title={Geographical knowledge-driven representation learning for remote sensing images},
  author={Li, Wenyuan and Chen, Keyan and Chen, Hao and Shi, Zhenwei},
  journal={IEEE Transactions on Geoscience and Remote Sensing},
  volume={60},
  pages={1--16},
  year={2021},
  publisher={IEEE}
}

@article{zhang2022consecutive,
  title={Consecutive pre-training: A knowledge transfer learning strategy with relevant unlabeled data for remote sensing domain},
  author={Zhang, Tong and Gao, Peng and Dong, Hao and Zhuang, Yin and Wang, Guanqun and Zhang, Wei and Chen, He},
  journal={Remote Sensing},
  volume={14},
  number={22},
  pages={5675},
  year={2022},
  publisher={MDPI}
}

@article{wang2024multi,
  title={Multi-Label Guided Soft Contrastive Learning for Efficient Earth Observation Pretraining},
  author={Wang, Yi and Albrecht, Conrad M and Zhu, Xiao Xiang},
  journal={arXiv preprint arXiv:2405.20462},
  year={2024}
}

@article{Bodnar2024a,
  title={A Foundation Model for the Earth System},
  author={Bodnar, Cristian and Bruinsma, Wessel P. and Lucic, Ana and Stanley, Megan and Vaughan, Anna and Brandstetter, Johannes and Garvan, Patrick and Riechert, Maik and Weyn, Jonathan A. and Dong, Haiyu and Gupta, Jayesh K. and Thambiratnam, Kit and Archibald, Alexander T. and Wu, Chun-Chieh and Heider, Elizabeth and Welling, Max and Turner, Richard E. and Perdikaris, Paris},
  journal={arXiv preprint arXiv:2405.13063},
  year={2024}
}

@article{dou2024cc2vec,
  title={CC2Vec: Combining Typed Tokens with Contrastive Learning for Effective Code Clone Detection},
  author={Dou, Shihan and Wu, Yueming and Jia, Haoxiang and Zhou, Yuhao and Liu, Yan and Liu, Yang},
  journal={Proceedings of the ACM on Software Engineering},
  volume={1},
  number={FSE},
  pages={1564--1584},
  year={2024},
  publisher={ACM New York, NY, USA}
}

@article{chen2023foundation,
  title={Foundation models for weather and climate data understanding: A comprehensive survey},
  author={Chen, Shengchao and Long, Guodong and Jiang, Jing and Liu, Dikai and Zhang, Chengqi},
  journal={arXiv preprint arXiv:2312.03014},
  year={2023}
}

@article{zhu2024foundations,
  title={On the Foundations of Earth and Climate Foundation Models},
  author={Zhu, Xiao Xiang and Xiong, Zhitong and Wang, Yi and Stewart, Adam J and Heidler, Konrad and Wang, Yuanyuan and Yuan, Zhenghang and Dujardin, Thomas and Xu, Qingsong and Shi, Yilei},
  journal={arXiv preprint arXiv:2405.04285},
  year={2024}
}

@inproceedings{wang2024decoupling,
  title={Decoupling common and unique representations for multimodal self-supervised learning},
  author={Wang, Yi and Albrecht, Conrad M and Braham, Nassim Ait Ali and Liu, Chenying and Xiong, Zhitong and Zhu, Xiao Xiang},
  booktitle={European Conference on Computer Vision},
  pages={286--303},
  year={2024},
  organization={Springer}
}

@inproceedings{bastani2023satlaspretrain,
  title={Satlaspretrain: A large-scale dataset for remote sensing image understanding},
  author={Bastani, Favyen and Wolters, Piper and Gupta, Ritwik and Ferdinando, Joe and Kembhavi, Aniruddha},
  booktitle={Proceedings of the IEEE/CVF International Conference on Computer Vision},
  pages={16772--16782},
  year={2023}
}

@inproceedings{guo2024skysense,
  title={Skysense: A multi-modal remote sensing foundation model towards universal interpretation for earth observation imagery},
  author={Guo, Xin and Lao, Jiangwei and Dang, Bo and Zhang, Yingying and Yu, Lei and Ru, Lixiang and Zhong, Liheng and Huang, Ziyuan and Wu, Kang and Hu, Dingxiang and others},
  booktitle={Proceedings of the IEEE/CVF Conference on Computer Vision and Pattern Recognition},
  pages={27672--27683},
  year={2024}
}

@inproceedings{nedungadi2024mmearth,
  title={MMEarth: Exploring multi-modal pretext tasks for geospatial representation learning},
  author={Nedungadi, Vishal and Kariryaa, Ankit and Oehmcke, Stefan and Belongie, Serge and Igel, Christian and Lang, Nico},
  booktitle={European Conference on Computer Vision},
  pages={164--182},
  year={2024},
  organization={Springer}
}

@article{ravirathinam2024towards,
  title={Towards a Knowledge guided Multimodal Foundation Model for Spatio-Temporal Remote Sensing Applications},
  author={Ravirathinam, Praveen and Khandelwal, Ankush and Ghosh, Rahul and Kumar, Vipin},
  journal={arXiv preprint arXiv:2407.19660},
  year={2024}
}

@article{luo2024mmm,
  title={MMM-RS: A Multi-modal, Multi-GSD, Multi-scene Remote Sensing Dataset and Benchmark for Text-to-Image Generation},
  author={Luo, Jialin and Wang, Yuanzhi and Gu, Ziqi and Qiu, Yide and Yao, Shuaizhen and Wang, Fuyun and Xu, Chunyan and Zhang, Wenhua and Wang, Dan and Cui, Zhen},
  journal={arXiv preprint arXiv:2410.22362},
  year={2024}
}

@article{yan2023ringmo,
  title={RingMo-SAM: A foundation model for segment anything in multimodal remote-sensing images},
  author={Yan, Zhiyuan and Li, Junxi and Li, Xuexue and Zhou, Ruixue and Zhang, Wenkai and Feng, Yingchao and Diao, Wenhui and Fu, Kun and Sun, Xian},
  journal={IEEE Transactions on Geoscience and Remote Sensing},
  volume={61},
  pages={1--16},
  year={2023},
  publisher={IEEE}
}

@article{xiao2024segment,
  title={Segment anything with multiple modalities},
  author={Xiao, Aoran and Xuan, Weihao and Qi, Heli and Xing, Yun and Yokoya, Naoto and Lu, Shijian},
  journal={arXiv preprint arXiv:2408.09085},
  year={2024}
}

@article{zhang20242,
  title={A$^2$-MAE: A spatial-temporal-spectral unified remote sensing pre-training method based on anchor-aware masked autoencoder},
  author={Zhang, Lixian and Zhao, Yi and Dong, Runmin and Zhang, Jinxiao and Yuan, Shuai and Cao, Shilei and Chen, Mengxuan and Zheng, Juepeng and Li, Weijia and Liu, Wei and others},
  journal={arXiv preprint arXiv:2406.08079},
  year={2024}
}

@inproceedings{manas2021seasonal,
  title={Seasonal contrast: Unsupervised pre-training from uncurated remote sensing data},
  author={Manas, Oscar and Lacoste, Alexandre and Gir{\'o}-i-Nieto, Xavier and Vazquez, David and Rodriguez, Pau},
  booktitle={Proceedings of the IEEE/CVF International Conference on Computer Vision},
  pages={9414--9423},
  year={2021}
}

@inproceedings{mall2023change,
  title={Change-aware sampling and contrastive learning for satellite images},
  author={Mall, Utkarsh and Hariharan, Bharath and Bala, Kavita},
  booktitle={Proceedings of the IEEE/CVF Conference on Computer Vision and Pattern Recognition},
  pages={5261--5270},
  year={2023}
}

@article{yao2023ringmo,
  title={RingMo-sense: Remote sensing foundation model for spatiotemporal prediction via spatiotemporal evolution disentangling},
  author={Yao, Fanglong and Lu, Wanxuan and Yang, Heming and Xu, Liangyu and Liu, Chenglong and Hu, Leiyi and Yu, Hongfeng and Liu, Nayu and Deng, Chubo and Tang, Deke and others},
  journal={IEEE Transactions on Geoscience and Remote Sensing},
  year={2023},
  publisher={IEEE}
}

@article{szwarcman2024prithvi,
  title={Prithvi-EO-2.0: A Versatile Multi-Temporal Foundation Model for Earth Observation Applications},
  author={Szwarcman, Daniela and Roy, Sujit and Fraccaro, Paolo and Gislason, Orsteinn Eli and Blumenstiel, Benedikt and Ghosal, Rinki and de Oliveira, Pedro Henrique and Almeida, Joao Lucas de Sousa and Sedona, Rocco and Kang, Yanghui and others},
  journal={arXiv preprint arXiv:2412.02732},
  year={2024}
}

@article{dumeur2024self,
  title={Self-supervised spatio-temporal representation learning of Satellite Image Time Series},
  author={Dumeur, Iris and Valero, Silvia and Inglada, Jordi},
  journal={IEEE Journal of Selected Topics in Applied Earth Observations and Remote Sensing},
  year={2024},
  publisher={IEEE}
}

@article{zhan2025skyeyegpt,
  title={Skyeyegpt: Unifying remote sensing vision-language tasks via instruction tuning with large language model},
  author={Zhan, Yang and Xiong, Zhitong and Yuan, Yuan},
  journal={ISPRS Journal of Photogrammetry and Remote Sensing},
  volume={221},
  pages={64--77},
  year={2025},
  publisher={Elsevier}
}

@article{hu2025rsgpt,
  title={Rsgpt: A remote sensing vision language model and benchmark},
  author={Hu, Yuan and Yuan, Jianlong and Wen, Congcong and Lu, Xiaonan and Liu, Yu and Li, Xiang},
  journal={ISPRS Journal of Photogrammetry and Remote Sensing},
  volume={224},
  pages={272--286},
  year={2025},
  publisher={Elsevier}
}

@inproceedings{kuckreja2024geochat,
  title={Geochat: Grounded large vision-language model for remote sensing},
  author={Kuckreja, Kartik and Danish, Muhammad Sohail and Naseer, Muzammal and Das, Abhijit and Khan, Salman and Khan, Fahad Shahbaz},
  booktitle={Proceedings of the IEEE/CVF Conference on Computer Vision and Pattern Recognition},
  pages={27831--27840},
  year={2024}
}

@article{pang2024vhm,
  title={VHM: Versatile and Honest Vision Language Model for Remote Sensing Image Analysis},
  author={Pang, Chao and Weng, Xingxing and Wu, Jiang and Li, Jiayu and Liu, Yi and Sun, Jiaxing and Li, Weijia and Wang, Shuai and Feng, Litong and Xia, Gui-Song and others},
  journal={arXiv preprint arXiv:2403.20213},
  year={2024}
}

@article{luo2024skysensegpt,
  title={Skysensegpt: A fine-grained instruction tuning dataset and model for remote sensing vision-language understanding},
  author={Luo, Junwei and Pang, Zhen and Zhang, Yongjun and Wang, Tingzhu and Wang, Linlin and Dang, Bo and Lao, Jiangwei and Wang, Jian and Chen, Jingdong and Tan, Yihua and others},
  journal={arXiv preprint arXiv:2406.10100},
  year={2024}
}

@article{bazi2024rs,
  title={Rs-llava: A large vision-language model for joint captioning and question answering in remote sensing imagery},
  author={Bazi, Yakoub and Bashmal, Laila and Al Rahhal, Mohamad Mahmoud and Ricci, Riccardo and Melgani, Farid},
  journal={Remote Sensing},
  volume={16},
  number={9},
  pages={1477},
  year={2024},
  publisher={MDPI}
}

@article{xu2024rs,
  title={RS-Agent: Automating Remote Sensing Tasks through Intelligent Agents},
  author={Xu, Wenjia and Yu, Zijian and Wang, Yixu and Wang, Jiuniu and Peng, Mugen},
  journal={arXiv preprint arXiv:2406.07089},
  year={2024}
}

@article{liu2024change,
  title={Change-agent: Towards interactive comprehensive remote sensing change interpretation and analysis},
  author={Liu, Chenyang and Chen, Keyan and Zhang, Haotian and Qi, Zipeng and Zou, Zhengxia and Shi, Zhenwei},
  journal={IEEE Transactions on Geoscience and Remote Sensing},
  year={2024},
  publisher={IEEE}
}

@article{zhang2024earthgpt,
  title={Earthgpt: A universal multi-modal large language model for multi-sensor image comprehension in remote sensing domain},
  author={Zhang, Wei and Cai, Miaoxin and Zhang, Tong and Zhuang, Yin and Mao, Xuerui},
  journal={IEEE Transactions on Geoscience and Remote Sensing},
  year={2024},
  publisher={IEEE}
}

@inproceedings{muhtar2024lhrs,
  title={Lhrs-bot: Empowering remote sensing with vgi-enhanced large multimodal language model},
  author={Muhtar, Dilxat and Li, Zhenshi and Gu, Feng and Zhang, Xueliang and Xiao, Pengfeng},
  booktitle={European Conference on Computer Vision},
  pages={440--457},
  year={2024},
  organization={Springer}
}

@article{zhang2024earthmarker,
  title={EarthMarker: Visual Prompt Learning for Region-level and Point-level Remote Sensing Imagery Comprehension},
  author={Zhang, Wei and Cai, Miaoxin and Zhang, Tong and Li, Jun and Zhuang, Yin and Mao, Xuerui},
  journal={arXiv preprint arXiv:2407.13596},
  year={2024}
}

@inproceedings{irvin2025teochat,
title={{TEOC}hat: A Large Vision-Language Assistant for Temporal Earth Observation Data},
author={Jeremy Andrew Irvin and Emily Ruoyu Liu and Joyce C. Chen and Ines Dormoy and Jinyoung Kim and Samar Khanna and Zhuo Zheng and Stefano Ermon},
booktitle={The Thirteenth International Conference on Learning Representations},
year={2025},
url={https://openreview.net/forum?id=pZz0nOroGv}
}

@inproceedings{reed2023scale,
  title={Scale-mae: A scale-aware masked autoencoder for multiscale geospatial representation learning},
  author={Reed, Colorado J and Gupta, Ritwik and Li, Shufan and Brockman, Sarah and Funk, Christopher and Clipp, Brian and Keutzer, Kurt and Candido, Salvatore and Uyttendaele, Matt and Darrell, Trevor},
  booktitle={Proceedings of the IEEE/CVF International Conference on Computer Vision},
  pages={4088--4099},
  year={2023}
}

@inproceedings{noman2024rethinking,
  title={Rethinking transformers pre-training for multi-spectral satellite imagery},
  author={Noman, Mubashir and Naseer, Muzammal and Cholakkal, Hisham and Anwer, Rao Muhammad and Khan, Salman and Khan, Fahad Shahbaz},
  booktitle={Proceedings of the IEEE/CVF Conference on Computer Vision and Pattern Recognition},
  pages={27811--27819},
  year={2024}
}

@article{tang2023cross,
  title={Cross-scale mae: A tale of multiscale exploitation in remote sensing},
  author={Tang, Maofeng and Cozma, Andrei and Georgiou, Konstantinos and Qi, Hairong},
  journal={Advances in Neural Information Processing Systems},
  volume={36},
  pages={20054--20066},
  year={2023}
}

@inproceedings{li2024masked,
  title={Masked angle-aware autoencoder for remote sensing images},
  author={Li, Zhihao and Hou, Biao and Ma, Siteng and Wu, Zitong and Guo, Xianpeng and Ren, Bo and Jiao, Licheng},
  booktitle={European Conference on Computer Vision},
  pages={260--278},
  year={2024},
  organization={Springer}
}

@article{wang2024scaling,
  title={Scaling efficient masked autoencoder learning on large remote sensing dataset},
  author={Wang, Fengxiang and Wang, Hongzhen and Wang, Di and Guo, Zonghao and Zhong, Zhenyu and Lan, Long and Zhang, Jing and Liu, Zhiyuan and Sun, Maosong},
  journal={arXiv e-prints},
  pages={arXiv--2406},
  year={2024}
}

@article{zhang2023ctxmim,
  title={CtxMIM: Context-enhanced masked image modeling for remote sensing image understanding},
  author={Zhang, Mingming and Liu, Qingjie and Wang, Yunhong},
  journal={arXiv preprint arXiv:2310.00022},
  year={2023}
}

@article{muhtar2023cmid,
  title={Cmid: A unified self-supervised learning framework for remote sensing image understanding},
  author={Muhtar, Dilxat and Zhang, Xueliang and Xiao, Pengfeng and Li, Zhenshi and Gu, Feng},
  journal={IEEE Transactions on Geoscience and Remote Sensing},
  volume={61},
  pages={1--17},
  year={2023},
  publisher={IEEE}
}

@article{diao2024ringmo,
  title={RingMo-Aerial: An Aerial Remote Sensing Foundation Model With A Affine Transformation Contrastive Learning},
  author={Diao, Wenhui and Yu, Haichen and Kang, Kaiyue and Ling, Tong and Liu, Di and Feng, Yingchao and Bi, Hanbo and Ren, Libo and Li, Xuexue and Mao, Yongqiang and others},
  journal={arXiv preprint arXiv:2409.13366},
  year={2024}
}

@article{cha2024billion,
  title={A Billion-scale Foundation Model for Remote Sensing Images},
  author={Cha, Keumgang and Seo, Junghoon and Lee, Taekyung},
  journal={IEEE Journal of Selected Topics in Applied Earth Observations and Remote Sensing},
  year={2024},
  publisher={IEEE}
}

@inproceedings{dias2024oreole,
  title={OReole-FM: successes and challenges toward billion-parameter foundation models for high-resolution satellite imagery},
  author={Dias, Philipe and Tsaris, Aristeidis and Bowman, Jordan and Potnis, Abhishek and Arndt, Jacob and Yang, H Lexie and Lunga, Dalton},
  booktitle={Proceedings of the 32nd ACM International Conference on Advances in Geographic Information Systems},
  pages={597--600},
  year={2024}
}

@article{lu2025pattern,
  title={Pattern Integration and Enhancement Vision Transformer for Self-supervised Learning in Remote Sensing},
  author={Lu, Kaixuan and Zhang, Ruiqian and Huang, Xiao and Xie, Yuxing and Ning, Xiaogang and Zhang, Hanchao and Yuan, Mengke and Zhang, Pan and Wang, Tao and Liao, Tongkui},
  journal={IEEE Transactions on Geoscience and Remote Sensing},
  year={2025},
  publisher={IEEE}
}

@article{zhou2024mesam,
  title={Mesam: Multiscale enhanced segment anything model for optical remote sensing images},
  author={Zhou, Xichuan and Liang, Fu and Chen, Lihui and Liu, Haijun and Song, Qianqian and Vivone, Gemine and Chanussot, Jocelyn},
  journal={IEEE Transactions on Geoscience and Remote Sensing},
  year={2024},
  publisher={IEEE}
}

@article{zhang2024rsam,
  title={Rsam-seg: A sam-based approach with prior knowledge integration for remote sensing image semantic segmentation},
  author={Zhang, Jie and Yang, Xubing and Jiang, Rui and Shao, Wei and Zhang, Li},
  journal={arXiv preprint arXiv:2402.19004},
  year={2024}
}

@article{chen2024rsprompter,
  title={RSPrompter: Learning to prompt for remote sensing instance segmentation based on visual foundation model},
  author={Chen, Keyan and Liu, Chenyang and Chen, Hao and Zhang, Haotian and Li, Wenyuan and Zou, Zhengxia and Shi, Zhenwei},
  journal={IEEE Transactions on Geoscience and Remote Sensing},
  volume={62},
  pages={1--17},
  year={2024},
  publisher={IEEE}
}

@inproceedings{zhang2024uv,
  title={Uv-sam: Adapting segment anything model for urban village identification},
  author={Zhang, Xin and Liu, Yu and Lin, Yuming and Liao, Qingmin and Li, Yong},
  booktitle={Proceedings of the AAAI Conference on Artificial Intelligence},
  volume={38},
  number={20},
  pages={22520--22528},
  year={2024}
}

@article{zhang2024alps,
  title={ALPS: An auto-labeling and pre-training scheme for remote sensing segmentation with segment anything model},
  author={Zhang, Song and Wang, Qingzhong and Liu, Junyi and Xiong, Haoyi},
  journal={arXiv preprint arXiv:2406.10855},
  year={2024}
}

@article{hu2024rs,
  title={RS-vHeat: Heat Conduction Guided Efficient Remote Sensing Foundation Model},
  author={Hu, Huiyang and Wang, Peijin and Bi, Hanbo and Tong, Boyuan and Wang, Zhaozhi and Diao, Wenhui and Chang, Hao and Feng, Yingchao and Zhang, Ziqi and Ye, Qixiang and others},
  journal={arXiv preprint arXiv:2411.17984},
  year={2024}
}

@article{pang2024hsigene,
  title={Hsigene: A foundation model for hyperspectral image generation},
  author={Pang, Li and Cao, Xiangyong and Tang, Datao and Xu, Shuang and Bai, Xueru and Zhou, Feng and Meng, Deyu},
  journal={arXiv preprint arXiv:2409.12470},
  year={2024}
}

@article{gong2024crossearth,
  title={Crossearth: Geospatial vision foundation model for domain generalizable remote sensing semantic segmentation},
  author={Gong, Ziyang and Wei, Zhixiang and Wang, Di and Ma, Xianzheng and Chen, Hongruixuan and Jia, Yuru and Deng, Yupeng and Ji, Zhenming and Zhu, Xiangwei and Yokoya, Naoto and others},
  journal={arXiv preprint arXiv:2410.22629},
  year={2024}
}

@article{astruc2024anysat,
  title={AnySat: An Earth Observation Model for Any Resolutions, Scales, and Modalities},
  author={Astruc, Guillaume and Gonthier, Nicolas and Mallet, Clement and Landrieu, Loic},
  journal={arXiv preprint arXiv:2412.14123},
  year={2024}
}

@article{yu2024metaearth,
  title={Metaearth: A generative foundation model for global-scale remote sensing image generation},
  author={Yu, Zhiping and Liu, Chenyang and Liu, Liqin and Shi, Zhenwei and Zou, Zhengxia},
  journal={IEEE Transactions on Pattern Analysis and Machine Intelligence},
  year={2024},
  publisher={IEEE}
}

@inproceedings{chu2024towards,
  title={Towards natural language-guided drones: GeoText-1652 benchmark with spatial relation matching},
  author={Chu, Meng and Zheng, Zhedong and Ji, Wei and Wang, Tingyu and Chua, Tat-Seng},
  booktitle={European Conference on Computer Vision},
  pages={213--231},
  year={2024},
  organization={Springer}
}

@article{liu2025text2earth,
  title={Text2Earth: Unlocking Text-driven Remote Sensing Image Generation with a Global-Scale Dataset and a Foundation Model},
  author={Liu, Chenyang and Chen, Keyan and Zhao, Rui and Zou, Zhengxia and Shi, Zhenwei},
  journal={arXiv preprint arXiv:2501.00895},
  year={2025}
}

@inproceedings{khanna2024diffusionsat,
title={DiffusionSat: A Generative Foundation Model for Satellite Imagery},
author={Samar Khanna and Patrick Liu and Linqi Zhou and Chenlin Meng and Robin Rombach and Marshall Burke and David B. Lobell and Stefano Ermon},
booktitle={The Twelfth International Conference on Learning Representations},
year={2024}
}

@article{tang2403crs,
  title={Crs-diff: Controllable generative remote sensing foundation model. arXiv 2024},
  author={Tang, D and Cao, X and Hou, X and Jiang, Z and Meng, D},
  journal={arXiv preprint arXiv:2403.11614}
}

@article{li2024lhrs,
  title={Lhrs-bot-nova: Improved multimodal large language model for remote sensing vision-language interpretation},
  author={Li, Zhenshi and Muhtar, Dilxat and Gu, Feng and Zhang, Xueliang and Xiao, Pengfeng and He, Guangjun and Zhu, Xiaoxiang},
  journal={arXiv preprint arXiv:2411.09301},
  year={2024}
}

@article{soni2024earthdial,
  title={Earthdial: Turning multi-sensory earth observations to interactive dialogues},
  author={Soni, Sagar and Dudhane, Akshay and Debary, Hiyam and Fiaz, Mustansar and Munir, Muhammad Akhtar and Danish, Muhammad Sohail and Fraccaro, Paolo and Watson, Campbell D and Klein, Levente J and Khan, Fahad Shahbaz and others},
  journal={arXiv preprint arXiv:2412.15190},
  year={2024}
}

@article{li2024unirs,
  title={UniRS: Unifying Multi-temporal Remote Sensing Tasks through Vision Language Models},
  author={Li, Yujie and Xu, Wenjia and Li, Guangzuo and Yu, Zijian and Wei, Zhiwei and Wang, Jiuniu and Peng, Mugen},
  journal={arXiv preprint arXiv:2412.20742},
  year={2024}
}

@article{xue2024reo,
  title={REO-VLM: Transforming VLM to Meet Regression Challenges in Earth Observation},
  author={Xue, Xizhe and Wei, Guoting and Chen, Hao and Zhang, Haokui and Lin, Feng and Shen, Chunhua and Zhu, Xiao Xiang},
  journal={arXiv preprint arXiv:2412.16583},
  year={2024}
}

@article{ou2025geopix,
  title={GeoPix: Multi-Modal Large Language Model for Pixel-level Image Understanding in Remote Sensing},
  author={Ou, Ruizhe and Hu, Yuan and Zhang, Fan and Chen, Jiaxin and Liu, Yu},
  journal={arXiv preprint arXiv:2501.06828},
  year={2025}
}

@article{shabbir2025geopixel,
  title={GeoPixel: Pixel Grounding Large Multimodal Model in Remote Sensing},
  author={Shabbir, Akashah and Zumri, Mohammed and Bennamoun, Mohammed and Khan, Fahad S and Khan, Salman},
  journal={arXiv preprint arXiv:2501.13925},
  year={2025}
}

@article{liu2024rsunivlm,
  title={RSUniVLM: A Unified Vision Language Model for Remote Sensing via Granularity-oriented Mixture of Experts},
  author={Liu, Xu and Lian, Zhouhui},
  journal={arXiv preprint arXiv:2412.05679},
  year={2024}
}

@article{wang2024ringmogpt,
  title={Ringmogpt: A unified remote sensing foundation model for vision, language, and grounded tasks},
  author={Wang, Peijin and Hu, Huiyang and Tong, Boyuan and Zhang, Ziqi and Yao, Fanglong and Feng, Yingchao and Zhu, Zining and Chang, Hao and Diao, Wenhui and Ye, Qixiang and others},
  journal={IEEE Transactions on Geoscience and Remote Sensing},
  year={2024},
  publisher={IEEE}
}

@article{huang2024generic,
  title={Generic knowledge boosted pretraining for remote sensing images},
  author={Huang, Ziyue and Zhang, Mingming and Gong, Yuan and Liu, Qingjie and Wang, Yunhong},
  journal={IEEE Transactions on Geoscience and Remote Sensing},
  volume={62},
  pages={1--13},
  year={2024},
  publisher={IEEE}
}

@inproceedings{prexl2023multi,
  title={Multi-modal multi-objective contrastive learning for sentinel-1/2 imagery},
  author={Prexl, Jonathan and Schmitt, Michael},
  booktitle={Proceedings of the IEEE/CVF Conference on Computer Vision and Pattern Recognition},
  pages={2136--2144},
  year={2023}
}

@inproceedings{wang2022self,
  title={Self-supervised vision transformers for joint SAR-optical representation learning},
  author={Wang, Yi and Albrecht, Conrad M and Zhu, Xiao Xiang},
  booktitle={IGARSS 2022-2022 IEEE International Geoscience and Remote Sensing Symposium},
  pages={139--142},
  year={2022},
  organization={IEEE}
}

@article{klemmer2023satclip,
  title={Satclip: Global, general-purpose location embeddings with satellite imagery},
  author={Klemmer, Konstantin and Rolf, Esther and Robinson, Caleb and Mackey, Lester and Ru{\ss}wurm, Marc},
  journal={arXiv preprint arXiv:2311.17179},
  year={2023}
}

@article{vivanco2023geoclip,
  title={Geoclip: Clip-inspired alignment between locations and images for effective worldwide geo-localization},
  author={Vivanco Cepeda, Vicente and Nayak, Gaurav Kumar and Shah, Mubarak},
  journal={Advances in Neural Information Processing Systems},
  volume={36},
  pages={8690--8701},
  year={2023}
}

@article{jiao2023brain,
  title={Brain-inspired remote sensing foundation models and open problems: A comprehensive survey},
  author={Jiao, Licheng and Huang, Zhongjian and Lu, Xiaoqiang and Liu, Xu and Yang, Yuting and Zhao, Jiaxuan and Zhang, Jinyue and Hou, Biao and Yang, Shuyuan and Liu, Fang and others},
  journal={IEEE Journal of Selected Topics in Applied Earth Observations and Remote Sensing},
  volume={16},
  pages={10084--10120},
  year={2023},
  publisher={IEEE}
}

@article{li2024vision,
  title={Vision-language models in remote sensing: Current progress and future trends},
  author={Li, Xiang and Wen, Congcong and Hu, Yuan and Yuan, Zhenghang and Zhu, Xiao Xiang},
  journal={IEEE Geoscience and Remote Sensing Magazine},
  year={2024},
  publisher={IEEE}
}

@article{tan2023promises,
  title={On the promises and challenges of multimodal foundation models for geographical, environmental, agricultural, and urban planning applications},
  author={Tan, Chenjiao and Cao, Qian and Li, Yiwei and Zhang, Jielu and Yang, Xiao and Zhao, Huaqin and Wu, Zihao and Liu, Zhengliang and Yang, Hao and Wu, Nemin and others},
  journal={arXiv preprint arXiv:2312.17016},
  year={2023}
}

@article{zhou2024towards,
  title={Towards vision-language geo-foundation model: A survey},
  author={Zhou, Yue and Feng, Litong and Ke, Yiping and Jiang, Xue and Yan, Junchi and Yang, Xue and Zhang, Wayne},
  journal={arXiv preprint arXiv:2406.09385},
  year={2024}
}

@article{lu2024ai,
  title={AI foundation models in remote sensing: A survey},
  author={Lu, Siqi and Guo, Junlin and Zimmer-Dauphinee, James R and Nieusma, Jordan M and Wang, Xiao and VanValkenburgh, Parker and Wernke, Steven A and Huo, Yuankai},
  journal={arXiv preprint arXiv:2408.03464},
  year={2024}
}

@article{zhang2024foundation,
  title={Foundation model for generalist remote sensing intelligence: Potentials and prospects},
  author={Zhang, Mi and Yang, Bingnan and Hu, Xiangyun and Gong, Jianya and Zhang, Zuxun},
  journal={Science Bulletin},
  volume={69},
  number={23},
  pages={3652--3656},
  year={2024},
  publisher={Elsevier}
}

@article{tao2025advancements,
  title={Advancements in Vision--Language Models for Remote Sensing: Datasets, Capabilities, and Enhancement Techniques},
  author={Tao, Lijie and Zhang, Haokui and Jing, Haizhao and Liu, Yu and Yan, Dawei and Wei, Guoting and Xue, Xizhe},
  journal={Remote Sensing},
  volume={17},
  number={1},
  pages={162},
  year={2025},
  publisher={MDPI}
}

@article{huo2025remote,
  title={When Remote Sensing Meets Foundation Model: A Survey and Beyond},
  author={Huo, Chunlei and Chen, Keming and Zhang, Shuaihao and Wang, Zeyu and Yan, Heyu and Shen, Jing and Hong, Yuyang and Qi, Geqi and Fang, Hongmei and Wang, Zihan},
  journal={remote sensing},
  volume={17},
  number={2},
  year={2025}
}

@article{xiao2024foundation,
  title={Foundation models for remote sensing and Earth Observation: A survey},
  author={Xiao, Aoran and Xuan, Weihao and Wang, Junjue and Huang, Jiaxing and Tao, Dacheng and Lu, Shijian and Yokoya, Naoto},
  journal={arXiv preprint arXiv:2410.16602},
  year={2024}
}

@article{zhang2024geoscience,
  title={When Geoscience Meets Foundation Models: Toward a general geoscience artificial intelligence system},
  author={Zhang, Hao and Xu, Jin-Jian and Cui, Hong-Wei and Li, Lin and Yang, Yaowen and Tang, Chao-Sheng and Boers, Niklas},
  journal={IEEE Geoscience and Remote Sensing Magazine},
  year={2024},
  publisher={IEEE}
}

@inproceedings{mai2023csp,
  title={Csp: Self-supervised contrastive spatial pre-training for geospatial-visual representations},
  author={Mai, Gengchen and Lao, Ni and He, Yutong and Song, Jiaming and Ermon, Stefano},
  booktitle={International Conference on Machine Learning},
  pages={23498--23515},
  year={2023},
  organization={PMLR}
}

@inproceedings{rolf2024mission,
  title={Mission Critical--Satellite Data is a Distinct Modality in Machine Learning},
  author={Rolf, Esther and Klemmer, Konstantin and Robinson, Caleb and Kerner, Hannah},
  booktitle={International Conference on Machine Learning},
  year={2024},
  organization={PMLR}
}

@inproceedings{sumbul2019bigearthnet,
  title={Bigearthnet: A large-scale benchmark archive for remote sensing image understanding},
  author={Sumbul, Gencer and Charfuelan, Marcela and Demir, Beg{\"u}m and Markl, Volker},
  booktitle={IGARSS 2019-2019 IEEE international geoscience and remote sensing symposium},
  pages={5901--5904},
  year={2019},
  organization={IEEE}
}

@article{hong2023cross,
  title={Cross-city matters: A multimodal remote sensing benchmark dataset for cross-city semantic segmentation using high-resolution domain adaptation networks},
  author={Hong, Danfeng and Zhang, Bing and Li, Hao and Li, Yuxuan and Yao, Jing and Li, Chenyu and Werner, Martin and Chanussot, Jocelyn and Zipf, Alexander and Zhu, Xiao Xiang},
  journal={Remote Sensing of Environment},
  volume={299},
  pages={113856},
  year={2023},
  publisher={Elsevier}
}

@article{schmitt2019sen12ms,
  title={SEN12MS--A curated dataset of georeferenced multi-spectral sentinel-1/2 imagery for deep learning and data fusion},
  author={Schmitt, Michael and Hughes, Lloyd Haydn and Qiu, Chunping and Zhu, Xiao Xiang},
  journal={arXiv preprint arXiv:1906.07789},
  year={2019}
}

@article{hong2020more,
  title={More diverse means better: Multimodal deep learning meets remote-sensing imagery classification},
  author={Hong, Danfeng and Gao, Lianru and Yokoya, Naoto and Yao, Jing and Chanussot, Jocelyn and Du, Qian and Zhang, Bing},
  journal={IEEE Transactions on Geoscience and Remote Sensing},
  volume={59},
  number={5},
  pages={4340--4354},
  year={2020},
  publisher={IEEE}
}

@article{vivone2024deep,
  title={Deep learning in remote sensing image fusion: Methods, protocols, data, and future perspectives},
  author={Vivone, Gemine and Deng, Liang-Jian and Deng, Shangqi and Hong, Danfeng and Jiang, Menghui and Li, Chenyu and Li, Wei and Shen, Huanfeng and Wu, Xiao and Xiao, Jin-Liang and others},
  journal={IEEE Geoscience and Remote Sensing Magazine},
  year={2024},
  publisher={IEEE}
}

@article{hackstein2024exploring,
  title={Exploring masked autoencoders for sensor-agnostic image retrieval in remote sensing},
  author={Hackstein, Jakob and Sumbul, Gencer and Clasen, Kai Norman and Demir, Beg{\"u}m},
  journal={IEEE Transactions on Geoscience and Remote Sensing},
  year={2024},
  publisher={IEEE}
}

@inproceedings{irvinteochat,
  title={TEOChat: A Large Vision-Language Assistant for Temporal Earth Observation Data},
  author={Irvin, Jeremy Andrew and Liu, Emily Ruoyu and Chen, Joyce C and Dormoy, Ines and Kim, Jinyoung and Khanna, Samar and Zheng, Zhuo and Ermon, Stefano},
  booktitle={The Thirteenth International Conference on Learning Representations},
  year={2025}
}

@article{hong2026hyperspectral,
  title={Hyperspectral Imaging},
  author={Hong, Danfeng and Li, Chenyu and Yokoya, Naoto and Zhang, Bing and Jia, Xiuping and Plaza, Antonio and Gamba, Paolo and Benediktsson, Jon Atli and Chanussot, Jocelyn},
  journal={Nature Reviews Methods Primers},
  year={2026}
}

@inproceedings{li2025reobench,
  title={REOBench: Benchmarking Robustness of Earth Observation Foundation Models},
  author={Li, Xiang and Tao, Yong and Zhang, Siyuan and Liu, Siwei and Xiong, Zhitong and Luo, Chunbo and Liu, Lu and Pechenizkiy, Mykola and Zhu, Xiao Xiang and Huang, Tianjin},
  booktitle={NeruIPS 2025 D\&B Track},
  year={2025}
}

@article{stewart2023ssl4eo,
  title={Ssl4eo-l: Datasets and foundation models for landsat imagery},
  author={Stewart, Adam and Lehmann, Nils and Corley, Isaac and Wang, Yi and Chang, Yi-Chia and Ait Ali Braham, Nassim Ait and Sehgal, Shradha and Robinson, Caleb and Banerjee, Arindam},
  journal={Advances in Neural Information Processing Systems},
  volume={36},
  pages={59787--59807},
  year={2023}
}

@article{wang2023ssl4eo,
  title={SSL4EO-S12: A large-scale multimodal, multitemporal dataset for self-supervised learning in Earth observation [Software and Data Sets]},
  author={Wang, Yi and Braham, Nassim Ait Ali and Xiong, Zhitong and Liu, Chenying and Albrecht, Conrad M and Zhu, Xiao Xiang},
  journal={IEEE Geoscience and Remote Sensing Magazine},
  volume={11},
  number={3},
  pages={98--106},
  year={2023},
  publisher={IEEE}
}

@article{li2023rs,
  title={RS-CLIP: Zero shot remote sensing scene classification via contrastive vision-language supervision},
  author={Li, Xiang and Wen, Congcong and Hu, Yuan and Zhou, Nan},
  journal={International Journal of Applied Earth Observation and Geoinformation},
  volume={124},
  pages={103497},
  year={2023},
  publisher={Elsevier}
}

\begin{IEEEbiography}[{\includegraphics[width=1in,height=1.25in,clip,keepaspectratio]{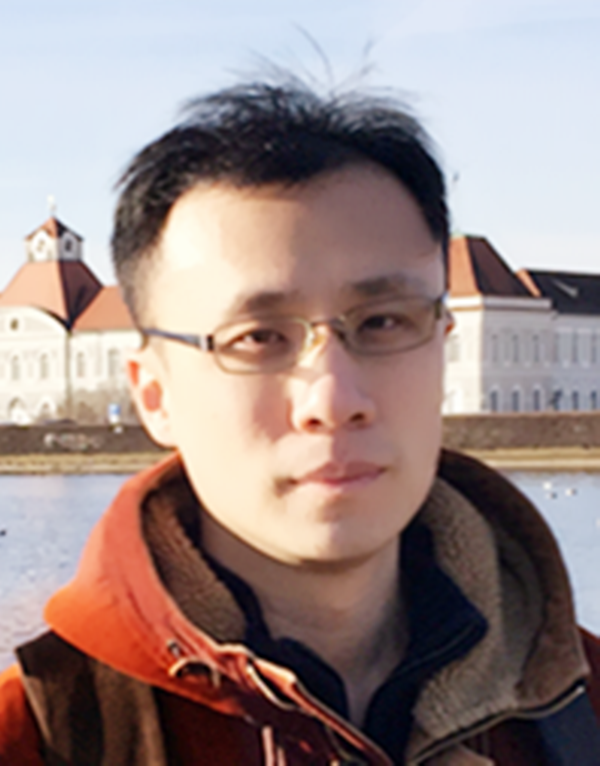}}]{Danfeng Hong}(IEEE Senior Member) received the Dr. -Ing degree (summa cum laude) from the Signal Processing in Earth Observation (SiPEO), Technical University of Munich (TUM), Munich, Germany, in 2019. He is currently a Full Professor at Southeast University, Nanjing, China, and previously served as a Full Professor at the Aerospace Information Research Institute, Chinese Academy of Sciences, Beijing, China. His research interests include artificial intelligence, multimodal perception, foundation models, hyperspectral remote sensing, and Earth science.

Dr. Hong serves as an Associate Editor for the IEEE Transactions on Pattern Analysis and Machine Intelligence (TPAMI), IEEE Transactions on Image Processing (TIP), and IEEE Transactions on Geoscience and Remote Sensing (TGRS). He has been recognized as a Highly Cited Researcher by Clarivate Analytics since 2022.
\end{IEEEbiography}

\vskip -2\baselineskip plus -1fil

\begin{IEEEbiography}[{\includegraphics[width=1in,height=1.25in,clip,keepaspectratio]
{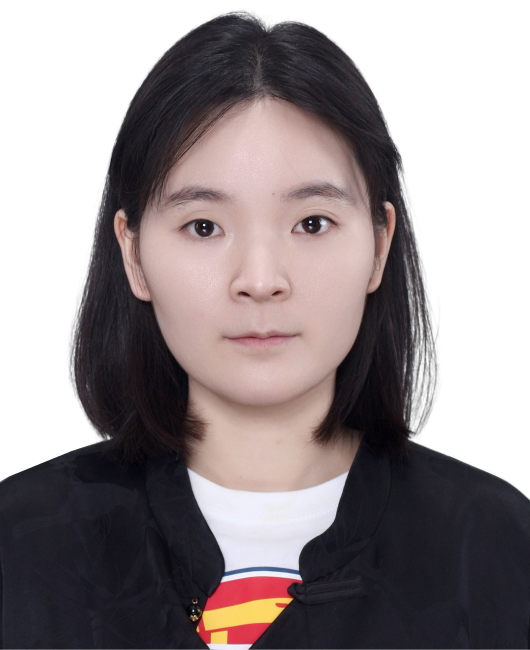}}]{Chenyu Li} received the B.S., M.S., and Ph.D. degrees from Southeast University, Nanjing, China, in 2018, 2021, and 2025, respectively. Her research interests include interpretable artificial intelligence, big Earth data forecasting, foundation models, and hyperspectral imaging.

Dr. Li received the outstanding Student Paper (Paul Gader) award at WHISPERS2025 and the Best Reviewer Award from IEEE Transactions on Geoscience and Remote Sensing (TGRS).
\end{IEEEbiography}

\vskip -2\baselineskip plus -1fil

\begin{IEEEbiography}[{\includegraphics[width=1in,height=1.25in,clip,keepaspectratio]{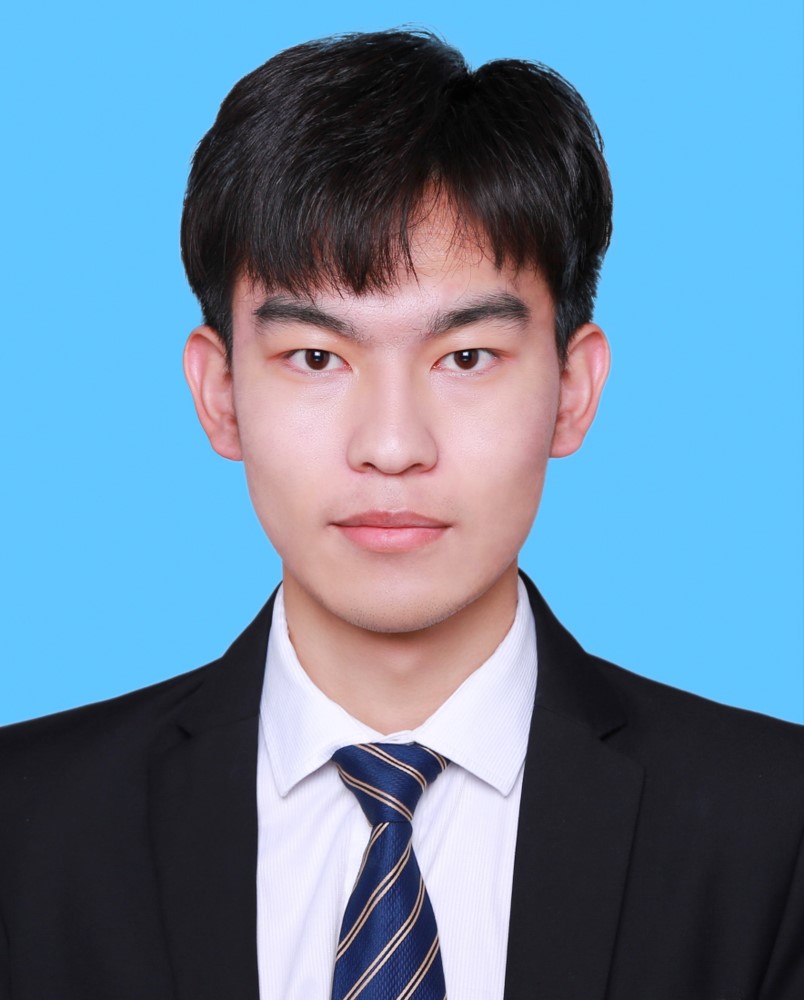}}]{Xuyang Li} received the B.S. degree from the School of Optoelectronic Science and Engineering, University of Electronic Science and Technology of China (UESTC), Chengdu, China, in 2023. He is currently pursuing his Ph.D. degree at the Aerospace Information Research Institute, Chinese Academy of Sciences, Beijing, China. His research interests include machine learning, remote sensing foundation models, and multimodal interpretation.
\end{IEEEbiography}

\vskip -2\baselineskip plus -1fil

\begin{IEEEbiography}[{\includegraphics[width=1in,height=1.25in,clip,keepaspectratio]{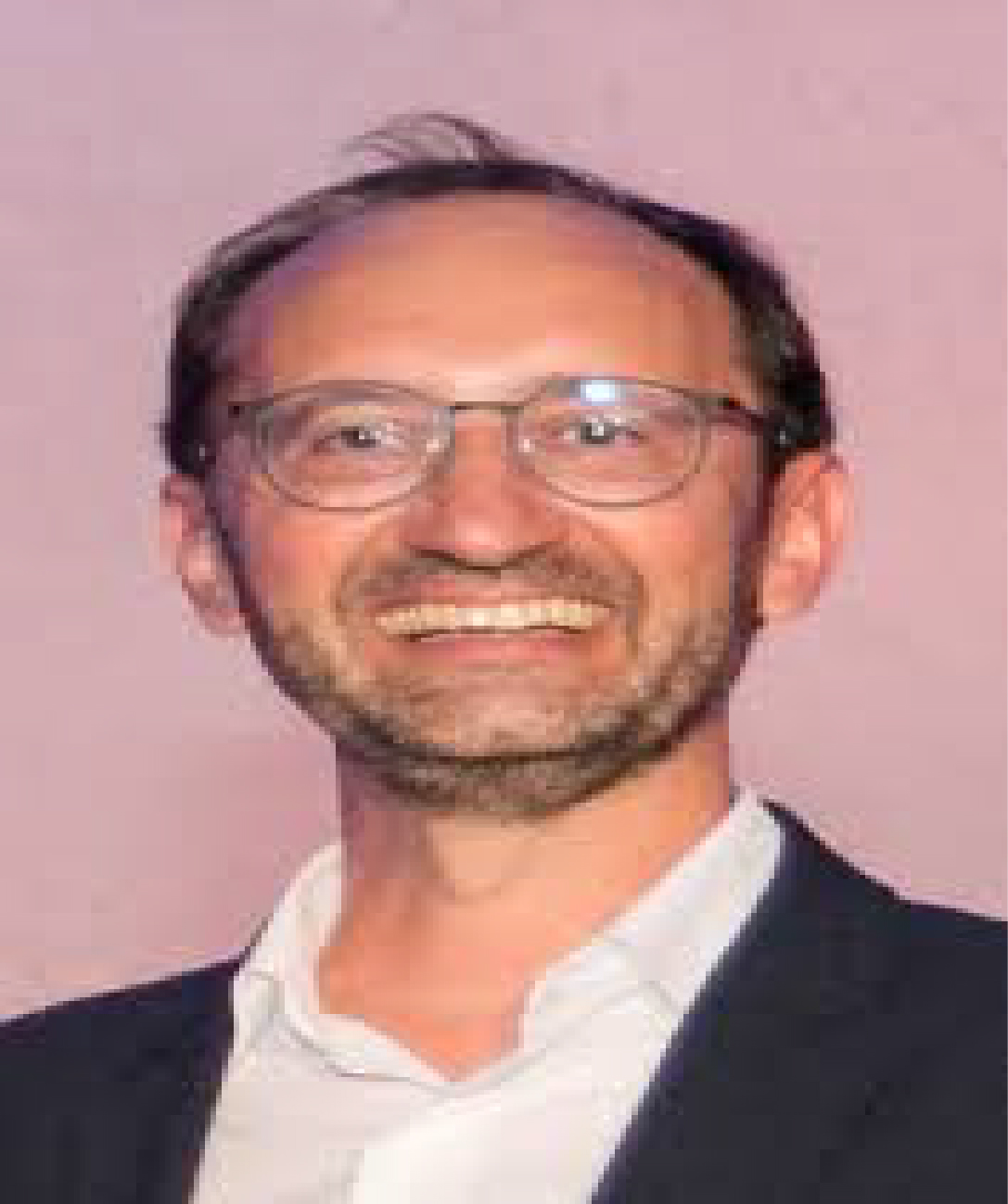}}]Gustau Camps-Valls (IEEE Fellow) is a Full Professor of Electrical Engineering at Universitat de València, Spain. He specializes in AI and machine learning for environmental and climate challenges. He heads the Image and Signal Processing (ISP) group, focusing on AI for Earth and climate sciences. His research encompasses AI for predicting extreme events, enhancing Earth models, and understanding climate impacts on society, such as climate-induced migration. He is a Highly Cited Researcher (2011, 2021-2024) and has received notable recognitions, including the Google Classic Paper Award (2019). He has been the Program Chair for IEEE IGARSS 2018 and AISTATS 2022, an Associate Editor for five IEEE journals, and an IEEE Distinguished Lecturer (2017-2019). He is a Fellow of IEEE, ELLIS, EurASc, Academia Europaea, and AAIA. His leadership extends to advisory roles, including at ESA PhiLab and EUMETSAT, and he coordinates ELLIS's AI program for Earth and climate sciences. He received two ERC grants (consolidator and synergy) to advance AI for Earth and climate sciences.
\end{IEEEbiography}

\vskip -2\baselineskip plus -1fil

\begin{IEEEbiography}[{\includegraphics[width=1in,height=1.25in,clip,keepaspectratio]{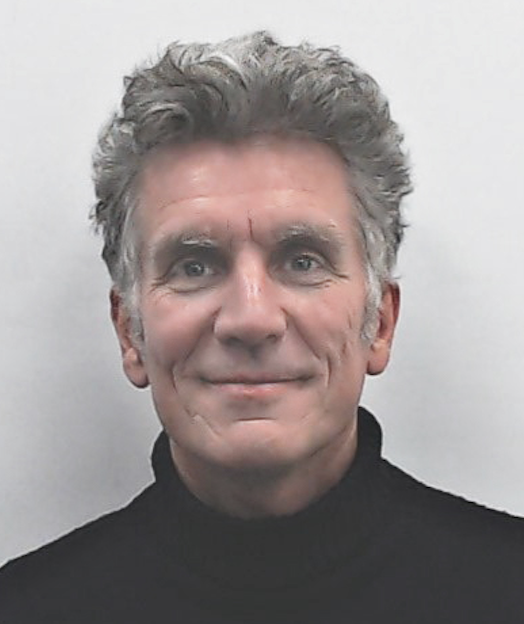}}]{Jocelyn Chanussot} (IEEE Fellow) received the M.Sc. degree in electrical engineering from the Grenoble Institute of Technology (Grenoble INP), Grenoble, France, in 1995, and the Ph.D. degree from the Université de Savoie, Annecy, France, in 1998. From 1999 to 2023, he was with Grenoble INP, where he was a Professor of signal and image processing. He is currently a Research Director with INRIA, Grenoble. His research interests include image analysis, hyperspectral remote sensing, data fusion, machine learning, and artificial intelligence.

Dr. Chanussot is the founding President of the IEEE Geoscience and Remote Sensing French chapter. He was the vice president of the IEEE Geoscience and Remote Sensing Society, in charge of meetings and symposia. He is an Associate Editor for the IEEE Transactions on Geoscience and Remote Sensing, the IEEE Transactions on Image Processing, and the Proceedings of the IEEE. He was the Editor-in-Chief of the IEEE Journal of Selected Topics in Applied Earth Observations and Remote Sensing (2011-2015). He is a Fellow of the IEEE, an ELLIS Fellow, a Fellow of AAIA, a member of the Institut Universitaire de France (2012-2017), and a Highly Cited Researcher (Clarivate Analytics/Thomson Reuters, since 2018).
\end{IEEEbiography}

\end{document}